\documentclass[a4paper,fleqn]{cas-sc}

\usepackage{caption} 
\usepackage{makecell} 
\usepackage{booktabs}
\usepackage{gensymb}
\usepackage{float}
\usepackage{amsmath}
\usepackage{amssymb}
\usepackage{xcolor}

\emergencystretch=\maxdimen
\usepackage[numbers]{natbib}

\def\tsc#1{\csdef{#1}{\textsc{\lowercase{#1}}\xspace}}
\tsc{WGM}
\tsc{QE}

\begin{document}
\let\WriteBookmarks\relax
\def\floatpagepagefraction{1}
\def\textpagefraction{.001}

\shorttitle{}    

\shortauthors{}  

\title [mode = title]{AdaPR: Adaptive Plane Reformatting for 4D Flow MRI using Deep Reinforcement Learning}  



\author[1,2,3]{Javier Bisbal}[orcid=0000-0002-1825-0097]
\cormark[1]
\ead{jebisbal@uc.cl}
\credit{Writing - Original Draft, Writing - Review \& Editing, Conceptualization, Methodology, Software, Validation, Formal analysis, Investigation, Data curation, Visualization, Project administration}

\author[4]{Julio Sotelo}[orcid=0000-0002-0915-5215]
\ead{julio.sotelo@usm.cl}
\credit{Writing - Review \& Editing, Methodology, Software, Validation, Formal analysis, Visualization}

\author[1]{Maria I Valdes}
\ead{mivaldesvar@gmail.com}
\credit{Data curation}

\author[1,2,3,5]{Pablo Irarrazaval}[orcid=0000-0002-5186-2642]
\ead{pim@uc.cl}
\credit{Writing – review \& editing}

\author[1,3,7]{Marcelo E Andia}[orcid=0000-0002-1251-5832]
\ead{meandia@uc.cl}
\credit{Writing – review \& editing}

\author[8]{Julio García}[orcid=0000-0001-9528-8155]
\ead{julio.garciaflores@ucalgary.ca}
\credit{Writing – review \& editing, Resources}

\author[9, 10,11,12]{José Rodriguez-Palomares}[orcid=0000-0002-7229-9780]
\ead{jfrodriguezpalomares@gmail.com}
\credit{Writing – review \& editing, Resources}

\author[13,14]{Francesca Raimondi}[orcid=0000-0003-2580-151X]
\ead{francesca.raimondi@gmail.com}
\credit{Writing – review \& editing, Resources}

\author[1,2,3]{Cristian Tejos}[orcid=0000-0002-8367-155X]
\ead{ctejos@uc.cl}
\credit{Writing – review \& editing, Supervision, Funding acquisition}

\author[6]{Sergio Uribe}[orcid=0000-0002-4970-9710]
\ead{sergio.uribe@monash.edu}
\credit{Writing – review \& editing, Conceptualization, Supervision, Funding acquisition}

\cortext[cor1]{Corresponding author: jebisbal@uc.cl}


\affiliation[1]{organization={Biomedical Imaging Center, Pontificia Universidad Católica de Chile},
                city={Santiago},
                country={Chile}}

\affiliation[2]{organization={Department of Electrical Engineering, School of Engineering, Pontificia Universidad Católica de Chile},
                city={Santiago},
                country={Chile}}

\affiliation[3]{organization={Millennium Institute for Intelligent Healthcare Engineering (iHEALTH)},
                city={Santiago},
                country={Chile}}

\affiliation[4]{organization={Departamento de Informática, Universidad Técnica Federico Santa María},
                city={Santiago},
                country={Chile}}

\affiliation[5]{organization={Institute for Biological and Medical Engineering, School of Engineering, Medicine and Biological Sciences, PUC},
                city={Santiago},
                country={Chile}}

\affiliation[6]{organization={Department of Medical Imaging and Radiation Sciences, Faculty of Medicine, Nursing and Health Sciences, Monash University},
                city={Melbourne},
                country={Australia}}

\affiliation[7]{organization={Department of Radiology, School of Medicine, PUC},
                city={Santiago},
                country={Chile}}

\affiliation[8]{organization={Cardiac Imaging Centre, Departments of Radiology and Cardiac Sciences, University of Calgary},
                city={Calgary, AB},
                country={Canada}}

\affiliation[9]{organization={Department of Cardiology, Vall d'Hebron Hospital Universitari, Vall d'Hebron Barcelona Hospital Campus},
                city={Barcelona},
                country={Spain}}

\affiliation[10]{organization={Cardiovascular Diseases, Vall d'Hebron Institut de Recerca (VHIR), Vall d'Hebron Barcelona Hospital Campus},
                city={Barcelona},
                country={Spain}}

\affiliation[11]{organization={Department of Medicine, Universitat Autònoma de Barcelona},
                city={Bellaterra},
                country={Spain}}
                
\affiliation[12]{organization={CIBER de Enfermedades Cardiovasculares, Instituto de Salud Carlos III},
                city={Madrid},
                country={Spain}}

\affiliation[13]{organization={Department of Cardiology and Cardiovascular Surgery, Papa Giovanni XXIII Hospital},
                city={Bergamo},
                country={Italy}}

\affiliation[14]{organization={Hôpital Necker Enfants Malades},
                city={Paris},
                country={France}}





\begin{abstract}
\noindent\textbf{Background and Objective:} Plane reformatting for Four-dimensional Phase Contrast Magnetic Resonance Imaging (4D flow MRI) is time-consuming and prone to inter-observer variability, which limits fast cardiovascular flow assessment. Deep reinforcement learning (DRL) trains agents to iteratively adjust plane position and orientation, enabling accurate plane reformatting without the need of detailed landmarks, making it suitable for images with limited contrast and resolution, such as 4D flow MRI. However, current DRL methods assume that test volumes share the same spatial alignment as training data, limiting generalization across scanners and institutions. To address this limitation, we introduce AdaPR (Adaptive Plane Reformatting), a DRL framework that uses a local coordinate system to navigate volumes with arbitrary positions and orientations.

\noindent\textbf{Methods:} We implemented AdaPR using the Asynchronous Advantage Actor–Critic (A3C) algorithm and validated on 88 4D flow MRI datasets acquired from multiple vendors, including patients with congenital heart disease.

\noindent\textbf{Results:} AdaPR achieved a mean angular error of 6.32° ± 4.15° and distance error of 3.40 ± 2.75 mm, outperforming global coordinate DRL methods and alternative non-DRL methods. AdaPR maintained consistent accuracy under different volume orientations and positions. Flow measurements from AdaPR planes showed no significant differences compared to two manual observers, with excellent correlation (R² = 0.972 and R² = 0.968), comparable to inter-observer agreement (R² = 0.969). 

\noindent\textbf{Conclusion:} AdaPR provides robust, orientation-independent plane reformatting for 4D flow MRI, achieving flow quantification comparable to expert observers. Its adaptability across datasets and scanners makes it a promising candidate for other medical imaging applications beyond 4D flow MRI.
\end{abstract}





\maketitle

\section{Introduction}
Four-dimensional Phase Contrast Magnetic Resonance Imaging, also known as 4D flow MRI, provides a time-resolved velocity field in a volumetric region of interest. 4D flow MRI quantifies advanced hemodynamic parameters for the assessment of cardiovascular diseases \citep{sotelo_impact_2022,raimondi_prevalence_2021,sotelo_fully_2022}. Although 4D flow MRI has facilitated the acquisition of the entire heart and great vessels \citep{uribe_four-dimensional_2009}, its analysis is complex and time-consuming, and laborious plane reformatting is required to retrospectively evaluate different flow quantities.

The positioning of planes within the volumetric region allows for specific hemodynamic analyzes of different vessels within the heart. However, orienting and positioning those planes in a double oblique manner is typically performed manually \citep{bissell_4d_2023}, making it user-dependent and time-consuming, especially when several vessels are analyzed, as in congenital heart diseases such as Tetralogy of Fallot, aortic coarctation, or bicuspid aortic valve \citep{zhong_intracardiac_2019, isorni_4d_2020}. 

Landmark localization has been a common approach for plane reformatting in cardiac MRI \citep{le2017computationally, blansit_deep_2019}. However, the relatively low spatial resolution and contrast of 4D flow MRI often prevent the precise identification of the landmarks necessary for accurate plane reformatting. Instead, techniques involving navigation within 3D Phase Contrast Magnetic Resonance Angiography (PC-MRA) and flow pathlines or streamlines have been adopted for manual plane reformatting \citep{bissell_4d_2023,rodriguez2018aortic}. 

Other approaches rely on atlas registration \cite{bustamante_atlas-based_2015,fan2025avp}. For 4D flow MRI, \citet{bustamante_atlas-based_2015} proposed a deformable registration between an atlas with reformatted planes and the input image, followed by flattening the reformatted planes using principal component analysis. More recently, for cardiac computed tomography (cardiac CT), \citet{fan2025avp} introduced a self-supervised, atlas-prompted strategy that maps arbitrary 2D slices to their corresponding 3D positions within a volume previously registered to a canonical atlas \citep{fan2025avp}. Although registration-based methods can standardize orientation and position for subsequent view planning or plane reformatting, their performance strongly depends on the similarity between volumes. Differences in resolution, field of view, or anatomy, especially those caused by heart malformations, can limit their generalizability.

Beyond landmark localization and registration-based techniques, deep learning techniques have been proposed to directly predict reformatted planes from volumetric data \cite{corrado_automatic_2022,alansary_automatic_2018,dou_agent_2019,huang_searching_2020}. For 4D flow MRI, \citet{corrado_automatic_2022} proposed a 3D convolutional neural network (CNN) to identify vessel patches and predict plane orientations and positions. However, its reliance on small vessel patches may overlook crucial information from surrounding structures. Furthermore, they reported flow differences within $[-1.88, 1.89]$ L/min limits of agreement, which correspond to approximately $35\%-40\%$ error for the ascending aorta and pulmonary artery.

Alternative deep learning approaches for direct plane reformatting are based on deep reinforcement learning (DRL) \citep{alansary_automatic_2018,dou_agent_2019,huang_searching_2020}. In this framework, a CNN-based agent sequentially predicts rotations and translations to update an initial plane towards an objective location. DRL obviates the need for landmarks; instead, it analyzes different structures in its navigation trajectory, making it particularly suitable for low-contrast images, such as 4D flow MRI.

Nevertheless, existing DRL algorithms for plane reformatting perform rotations and translations using a fixed global coordinate system defined by the MRI acquisition. Consequently, if the test volumes differ in orientation or position from the training data, the algorithm may fail to converge unless all volumes are pre-registered to a common spatial framework. This requirement significantly restricts the use of databases from different institutions or vendors. 

In this work, we present AdaPR (\textit{Adaptive Plane Reformatting}), a novel DRL algorithm for reformatting planes in 4D flow MRI data acquired in any orientation or position. Our approach employs a local coordinate system during plane transitions, thereby enabling flexibility across different orientations and positions, and improving both generalization and performance in automated plane reformatting.

\label{sec:introduction}
\section{Methods}

\subsection{Reinforcement learning framework}
In reinforcement learning, an agent learns to make decisions by interacting with an environment, receiving rewards, and refining its strategy to maximize the accumulated reward. 

In a reinforcement learning task, an agent sequentially updates a state $s_t \in S$ by performing action $a_t \in A$ at step $t \in [t_0, t_{\text{max}}]$. Here, $S$ is the set of possible states within an environment $E$, and $A$ represents the set of possible actions. For each transition $T(s_t, a_t, s_{t+1})$, i.e., for each action $a_t$ with which the agent transits from state $s_t$ to state $s_{t+1}$, we define the reward $r_t \in \mathcal{R}$, where $\mathcal{R}$ is the set of possible rewards. Finally, a termination signal is activated when the agent reaches a target state or exceeds $t_{\text{max}}$ steps. Upon receiving this signal, the environment is reset, and the agent starts a new sequence, often referred to as an episode.

For the task of plane reformatting in 4D flow MRI, we defined the following reinforcement learning framework:

\subsubsection{Environment ($E$)}
The environment in which the agent navigates considers two sets of volumes, 3D PC-MRA  \citep{hennemuth2011fast} and systolic velocities.
\\

\textbf{3D PC-MRA.} This volume was computed as:

\begin{equation}
    \text{3D PC-MRA} (\vec{r}) = \sqrt{\frac{1}{N} \sum_{\tau=1}^N {M_{\tau}}^2 (\vec{r}) \sum_{j=x,y,z} {V_{(\tau,j)}}^2 (\vec{r})},
    \label{eq:1}
\end{equation}
with $M_{\tau}$ the magnitude of the 4D flow MRI signal at timeframe $\tau$; $V_{(\tau,j)}$ the component $j$ of the velocity at timeframe $\tau$; $\vec{r}$ the three-dimensional spatial location; and $N$ the number of timeframes.

To reduce noise and enhance the contrast between vascular and non-vascular regions, we adapted Eq. \eqref{eq:1} to:
\begin{equation}
    \text{3D PC-MRA} (\vec{r}) = \sqrt{\frac{1}{N} \sum_{\tau=1}^N M_{\tau}^2 (\vec{r}) \sum_{j=x,y,z} (V_{(\tau,j)} (\vec{r}) K_{\tau} (\vec{r}))^2 },
    \label{eq:2}
\end{equation}
with,
\[
K_{\tau} (\vec{r}) = 1 - \frac{\log(\text{std}(|\nabla \cdot \vec{V}_{\tau} (\vec{r})|) )}{\max[\log(\text{std}(|\nabla \cdot \vec{V}_{\tau} (\vec{r})|) )]},
\]
where $\vec{V}_{\tau}$ represents the velocity vector at timeframe $\tau$ for the spatial location $\vec{r}$, $\nabla \cdot$ is the divergence operation in each spatial location $\vec{r}$, std() represents the standard deviation along the time dimension, and max() the maximum across all spatial dimensions. In simple terms, we multiplied the velocity by values close to 1 where the standard deviation of the divergence was low (mass conservation) and by values close to 0 where the standard deviation of the divergence was high (background noise).

Finally, to increase homogeneity between different 3D PC-MRA scans, we performed Contrast Limited Adaptive Histogram Equalization (CLAHE) on the result of equation \eqref{eq:2}. Unlike standard histogram equalization, CLAHE avoids noise amplification. It works with two parameters: contextual region size and clipping limit. For our 4D flow MRI datasets, we chose a contextual region of one sixth of the original image size and a clipping limit of $10^{-4}$ \citep{pizer_adaptive_1987}.\\

\textbf{Systolic velocities.} To provide more information to the agent, we extracted velocities at peak systole. We applied a threshold to the 3D PC-MRA image, ensuring the evaluation of the peak timeframe only in the regions of the vessels.

To decrease variability due to different resolutions, we scaled all data sets to an isotropic resolution of 2x2x2 mm$^3$ using trilinear interpolation. Following the reformatting process, this scaling was then reversed before the flow computation.

\subsubsection{States ($S$)}\label{sec:states}

We define the agent's state in two ways, local (AdaPR) and global (VanillaPR), to contrast our adaptive formulation with a baseline.

\textbf{Local coordinates (AdaPR):} To allow our method to work with volumes in different orientations or positions, each state of AdaPR at step $t$ is defined by a central point $\vec{P}_t=P_t^x \hat{\imath}+P_t^y \hat{\jmath}+P_t^z \hat{k}$ and a local orthonormal basis $B_t=\{\vec{n}_t, \vec{w}_t^1, \vec{w}_t^2\}$. Each state is then represented by an array of 3D sub-coordinates centered at $\vec{P}_t$, with the first, second and third axes oriented along the $\vec{n}_t$, $\vec{w}_t^1$, and $\vec{w}_t^2$ vector directions, respectively. The sub-coordinates have dimensions $(D,H,W)$, where $D$, $H$ and $W$ correspond to depth, height, and width, respectively, with 2x2x2 mm$^3$ spacing. The reformatted plane at step $t$ is then defined by the middle slice along the first (depth) dimension, i.e., the slice at index $D/2$.


Because different bases $B_t$ can produce identical slice content in different orientations, we defined auxiliary lines that go through the center of each slice along the first dimension, following the directions of $\vec{w}_t^1$ and $\vec{w}_t^2$. These lines preserve orientation information even when the image content is the same.

\textbf{Global coordinates (VanillaPR)}: "VanillaPR" denotes the baseline framework of Alansary et al. \citep{alansary_automatic_2018}. Here, the reformatted plane is defined by the implicit equation of the plane $ax+by+cz+d=0$, where $(a, b, c)$ is the unit normal vector to the plane and $d$ is its signed distance from the origin (center of the volume). We generate a regular subgrid of size $(D,H,W)$ centered on the origin of the volume, then rotate and translate this grid according to $(a, b, c,d)$.

Using either the local or global sub-coordinate grids (AdaPR or VanillaPR), we sampled two subvolumes via trilinear interpolation: the first from the 3D PC-MRA volume and the second containing the projections of the systolic velocities
\begin{equation}
V_t = \bigl\lvert \vec{n}_t \cdot \vec{V}_{\mathrm{sys}} \bigr\rvert,
\quad
\vec{V}_{\mathrm{sys}} = V_{\mathrm{sys},x}^x\,\hat{\imath}
                      + V_{\mathrm{sys}}^y\,\hat{\jmath}
                      + V_{\mathrm{sys}}^z\,\hat{k}.
\end{equation}
where “sys” denotes the peak‐systolic timeframe.


\textbf{Initial state ($s_0$):} During training, the initial center $\vec{P}_0$ is sampled uniformly within 10\% of the volume size around the center and the initial normal $\vec{n}_0$ is drawn uniformly from the unit sphere. For the formulation of local coordinates, $\vec{w}_0^1$ is randomly chosen from all unit vectors orthogonal to $\vec{n}_0$, and $\vec{w}_0^2 =\vec{n}_0 \times \vec{w}_0^1  $.

\subsubsection{Actions ($A$)}

\noindent \textbf{Local coordinates (AdaPR):} For state $s_t$ with basis $\{\vec{n},\vec{w}_1,\vec{w}_2\}_t$ the set of actions is defined as:
\begin{equation}
A_t^{local}=\{(r_{\vec{w}_t^1},r_{\vec{w}_t^2},m_{\vec{w}_t^1},m_{\vec{w}_t^2},m_{\vec{n}_t}) \mid s_t \in S\},
\label{eq:actions_local}
\end{equation}
$r_{\vec{w}_t^1}$ denotes a rotation of $\vec{n}_t$ and $\vec{w}_t^2$ around the axis defined by $\vec{w}_t^1$; $r_{\vec{w}_t^2}$ denotes a rotation of $\vec{n}_t$ and $\vec{w}_t^1$ around the axis defined by $\vec{w}_t^2$; and $m_{\vec{w}_t^1},m_{\vec{w}_t^2},m_{\vec{n}_t}$ denote translations of $(\vec{P}_t)$ in the directions defined by $\vec{w}_t^1, \vec{w}_t^2$, and $\vec{n}_t$, respectively.

 Rotations are restricted to values between $[-\omega_{\text{max}}, \omega_{\text{max}}]$ degrees ($^\circ$) and translations to $[-d_{\text{max}}, d_{\text{max}}]$ millimeters (mm). These maximum values are discussed later.

\noindent \textbf{Global coordinates (VanillaPR):} For state $s_t$ the set of actions is defined as 
\begin{equation}
A_t^{global}=\{(a_{\theta_x},a_{\theta_y},a_{\theta_z},a_{d}) \mid s_t \in S\},
\label{eq:actions_global_1}
\end{equation}
where each action updates the plane parameters as 
\begin{equation}
\begin{aligned}
  a &\leftarrow \cos\bigl(\theta_x + a_{\theta_x}\bigr), &
  b &\leftarrow \cos\bigl(\theta_y + a_{\theta_y}\bigr), \\
  c &\leftarrow \cos\bigl(\theta_z + a_{\theta_z}\bigr), &
  d &\leftarrow d + a_d,
\end{aligned}
\label{eq:actions_global_2}
\end{equation}
with $\theta_i = \arccos(j)$ for $i\in\{x,y,z\}$ and $j\in\{a,b,c\}$. In this formulation, the parameters of $A_t^{global}$ are fixed for each transition but are adaptively reduced when oscillations occur near the target, following \citet{alansary_automatic_2018}.


\subsubsection{Reward ($\mathcal{R}$)}
We used two type of rewards, continuous-valued and discrete-valued, to evaluate both discrete and continuous DRL algorithms (\autoref{RL-algorithms}).

\textbf{Continuous-valued reward.} We defined the following cost function, which includes angular and distance errors:
\begin{equation}
C(t)=\left(1-\frac{\vec{n}_t \cdot \vec{n}_T}{\left\|\vec{n}_t\right\| \left\|\vec{n}_T\right\|}\right)+\lambda d(P_t,P_T),
\label{eq:cost_function}
\end{equation}
where $\vec{n}_T$ and $P_T$ are the normal vector and center point defining the target reformatted plane, respectively. $d(P_t,P_T)$ is the Euclidean distance between $P_t$ and $P_T$, $\lambda$ is a scalar that balances angular vs. distance error, and $\|\cdot\|$ denotes the Euclidean norm. We aim to find the set of transitions $T(s_t,a_t,s_{t+1})$ that minimize $C(t)$.

For each transition, we defined the reward as a discrete approximation of the temporal derivative of the cost function:
\begin{equation}
r_t=-\frac{dC}{dt}=C(t-1)-C(t).
\label{eq:reward}
\end{equation}
Following the negative direction of the gradient of $C(t)$ yields positive rewards until the agent reaches a state close to the target.

\textbf{Discrete-valued reward}: \citet{alansary_automatic_2018} defined a discrete reward as:

\begin{equation}
    r_t = \text{sign}\left(D(s_{t-1},s_T) - D(s_{t},s_T)\right)
\end{equation}

\noindent
where $D$ is a function that computes the Euclidean distance between the plane parameters ${a,b,c,d}$ of the current or previous state ($s_t$, $s_{t-1}$) and those of the target state $s_T$. In this formulation, the reward can take only three possible values $\{-1,0,1\}$.

\subsubsection{Terminal state}

The agent maximizes the expected return defined as $R_t=\sum_{k=0}^\infty \gamma^k r_{t+k}$, with $0<\gamma<1$. In this context, the agent maximizes $R_t$ with rewards approaching zero as it gets closer to the target. In the case of discrete-valued rewards, this behavior manifests as oscillations around the target. This behavior may slow the agent's convergence to the target plane location. To address this problem, we define a “near optimal” terminal state when the angle between $\vec{n}_t$ and $\vec{n}_T$ is less than 3$^\circ$ and the Euclidean distance between $P_t$ and $P_T$ is less than 2 mm. When these conditions are met, the agent receives an additional reward of 3.

We also defined a terminal state when the current step exceeded $t_{\text{max}}$. In this case, no extra reward is given to the agent. Upon reaching a terminal state, an initial state is sampled from a new 4D flow MRI dataset, beginning the next episode.

\begin{figure*}[!ht]
    \centering
    \includegraphics[width=0.8\textwidth,]{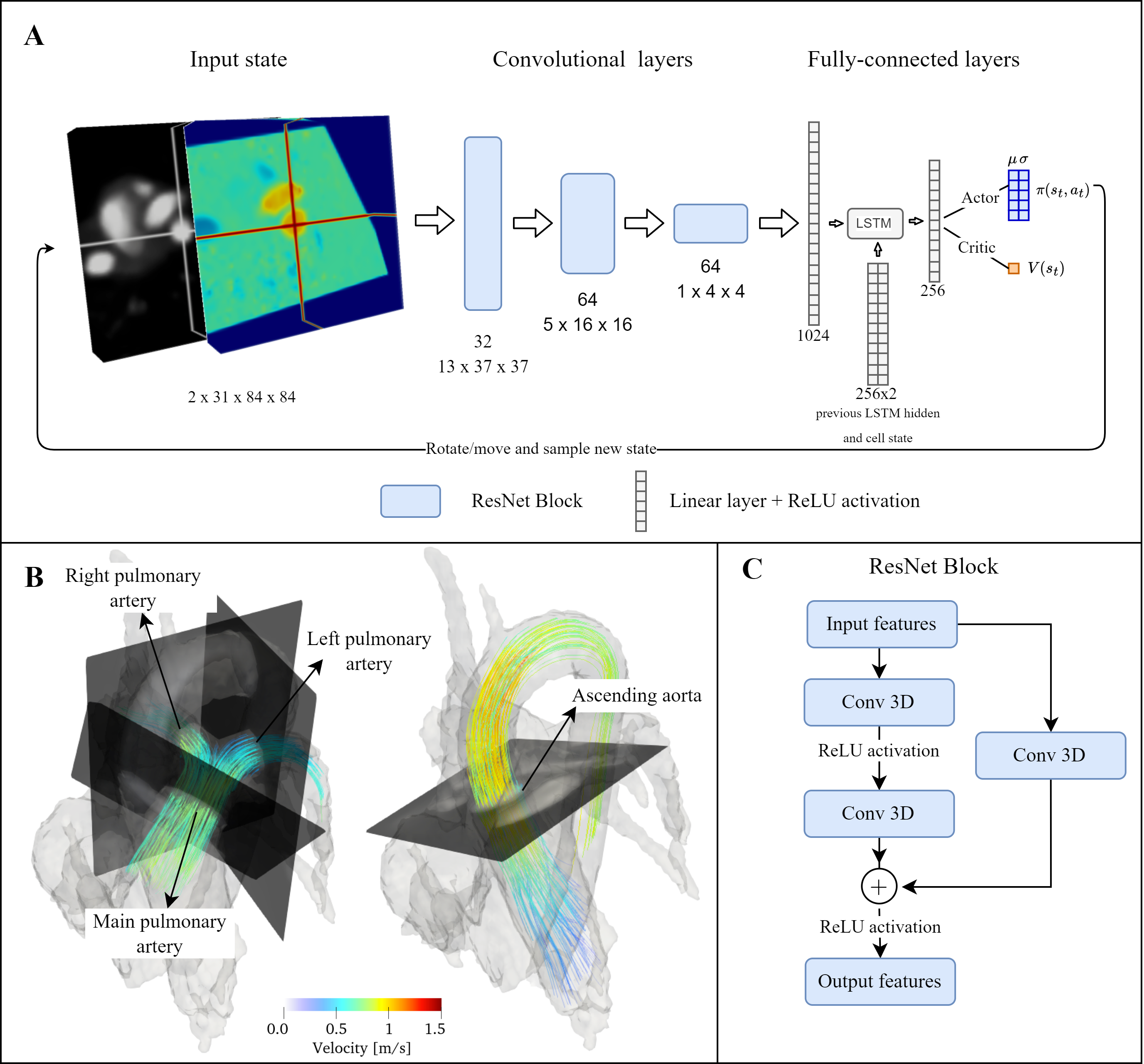}
    \caption{ \textbf{Overview of the adaptive plane-reformatting framework (AdaPR) and reformatted planes.} \textbf{A}. Asynchronous Advantage Actor–Critic (A3C) architecture:
    An initial stack of volumes goes through three 3D convolutional layers and 1 fully-connected layer. The output of this layer feeds an LSTM cell with the previous LSTM hidden and cell state. Next, the Actor layers compute $\mu_a$ and $\sigma_a$ to estimate the policy $\pi(s,a)$ and the Critic layer estimates the value function $G(s)$. The model parameters are updated with equation (6) and the policy is sampled to transition to a new state.
    \textbf{B}. Example of manual reformatted planes used for training.
    \textbf{C}. Diagram of ResNet blocks in convolutional layers.}
    \label{fig:Methods_overview}
\end{figure*}

\subsection{Reinforcement learning algorithms}\label{RL-algorithms}

Deep Q-Network (DQN), an off-policy deep reinforcement learning (DRL) algorithm, has been applied to plane reformatting using past experiences for batch learning \citep{alansary_automatic_2018}. In our framework, these experiences span diverse positions, orientations, and local coordinates. However, this heterogeneity in the state space can hinder learning by introducing instability and complexity to off-policy algorithms.

In contrast, on-policy algorithms focus on refining a single policy through active exploration, offering improved stability and consistency \citep{grondman_survey_2012}. To leverage these benefits, we implemented AdaPR using the Asynchronous Advantage Actor-Critic (A3C) algorithm \citep{mnih_asynchronous_2016}, hereafter referred to as A3C-AdaPR when compared to other DRL variants. 

For both DQN and A3C the agent is modeled as a CNN. The input of the CNN is the current state, structured as a 2-channel array of size $(2,D,H,W)$. The first channel contains the 3D PC-MRA sub-volume, while the second contains $V_t$. This array goes through four 3D convolutional layers with residual connections (ResNet) \citep{he_deep_2016}, creating a $1024 \times 1$ latent space representation. For DQN, this representation goes through a fully connected layer to estimate $Q(s_t,a_t)$, i.e., the expected return of every discrete action ($a_t$) in state $s_t$. For more details on DQN training, refer to \cite{alansary_automatic_2018}. 

For A3C, the agent estimates a policy $\pi(s,a)$, represented as a normal probability distribution parameterized by the expected value $\mu_a$ for each action and its standard deviation $\sigma_a$. In addition, the agent estimates a value function $G(s)=\mathbb{E}[R_t \mid s_t=s]$, indicating the expected return at state $s_t=s$. The CNN latent representation is fed into a Long Short-Term Memory (LSTM) cell \citep{hochreiter_long_1997}, which connects the current state representation to the previous state representation to exploit information between consecutive states. The LSTM output feeds into three parallel fully connected branches: two actor layers and one critic layer. The first actor layer uses a softsign activation to estimate the action mean $\mu$, while the second actor layer uses a softplus activation to estimate the standard deviation $\sigma$. The critic layer computes the state-value function $G(s)$ directly. The network was trained with the following gradient:
\begin{equation}
\nabla\log\pi (a_t \mid s_t )F(s_t,a_t )+\eta\nabla H(\pi(s_t )),
\label{eq:gradient}
\end{equation}
where $\nabla[\log\pi (a_t \mid s_t )]$ denotes the direction of the gradient that increases the probability of action $a_t$ in state $s_t$.

The advantage function $F(s_t,a_t)$ enforces the exploitation of the action policy, i.e., it estimates whether $a_t$ leads to an improvement over the expected return after $k_A$ steps, and is defined as,
\begin{equation}
F(s_t,a_t )=\sum_{i=0}^{k_A-1}\gamma^i r_{t+i} + \gamma^{k_A} G(s_{t+k_A} )-G(s_t ).
\label{eq:advantage}
\end{equation}
The entropy $H(\pi(s_t))$ enforces exploration of the action policy, i.e., it prevents premature convergence to suboptimal policies. For a normal distribution, it is defined as:
\begin{equation}
H(\pi(s_t))= \sum_{a \in A}\frac{1}{2}(\log_e(2\pi\sigma_a^2)+1).
\label{eq:entropy}
\end{equation}
The coefficient $\eta$ in equation \eqref{eq:gradient} balances exploration versus exploitation.

A diagram of the A3C reinforcement learning framework is shown in \autoref{fig:Methods_overview}.

\subsection{Experimental Setup}
\subsubsection{4D flow MRI data}

We processed a total of 88 anonymized datasets obtained with scanners from different vendors and different institutions, including healthy volunteers and patients with congenital heart defects (47 men, 34 $\pm$ 12.4 years old) (see \autoref{tab:4D_Flow_MRI_Datasets}). The patient cohort included individuals with conditions such as coarctation of the aorta (CoA), tetralogy of Fallot (TOF), and bicuspid aortic valve (BAV). These patients represented a wide range of ages and body sizes, but had otherwise normal cardiac anatomy. The authorization of each institution was obtained in accordance with local legislation and ethical compliance. Additional details on the scanning parameters for each dataset can be found in \autoref{Appendix}.

\begin{table}[bht]
\centering
\caption{4D flow MRI datasets}
\begin{tabular}{@{} p{1.7cm} p{1.15cm} l @{} }
\hline
\textbf{MRI scanner} & \textbf{Number of scans} & \textbf{Description} \\
\hline
GE 1.5T & 32 & Healthy volunteers \\
GE 1.5T & 8 & Patients with repaired aortic CoA \\
GE 1.5T & 5 & Patients with TOF \\
Siemens 3T & 17 & Healthy volunteers \\
Siemens 3T & 8 & Patients with BAV \\
Philips 3T & 18 & Healthy volunteers \\
\hline
\end{tabular}
\label{tab:4D_Flow_MRI_Datasets}
\end{table}

To evaluate plane reformatting performance across all scans, we performed four-fold cross-validation, stratified for each dataset (\autoref{tab:4D_Flow_MRI_Datasets}). For each cross-validation iteration, a blank model was trained with 52 scans ($\sim$59\%) of each dataset, validated with 14 ($\sim$16\%), and tested on the remaining 22 (25\%). This means that four models were trained to test 25\% of the data per iteration.

\subsubsection{Manually placed planes}

For each scan, we used ParaView software (Kitware, Clifton Park, NY 12065, USA) to generate 3D contours of the vessel’s walls using the 3D PC-MRA volume. We also generated velocity vectors within the contours. Two observers (O1 and O2) experienced in 4D flow MRI used the contours and velocity vectors to place planes in a double oblique orientation \citep{rodriguez2018aortic} in 4 vessels: pulmonary artery (PA), right pulmonary artery (RPA), left pulmonary artery (LPA), and ascending aorta (AAo). They also located a point at the center of the vessels to define $P_T$ as the target point of the automated plane reformatting. We used the planes placed by the first observer to train the DRL algorithms and the planes placed by the second observer to assess inter-observer variability.

\subsubsection{Alternative DRL methods}

To demonstrate the advantages of our proposed adaptive framework, we compared AdaPR against two alternatives:

\begin{itemize}
    \item DQN-VanillaPR \citep{alansary_automatic_2018}: The original DQN implementation for plane reformatting using the VanillaPR framework, with discrete actions fixed to a global coordinate system.
    \item A3C-VanillaPR: The A3C implementation proposed in this paper, but using the VanillaPR framework.   
\end{itemize}

Both alternatives were trained on data that was pre-aligned to the same orientation and position via rigid registration using the MATLAB Image Processing Toolbox (The MathWorks, Natick, MA). As a reference image for registration, we used the 3D PC-MRA of a healthy volunteer.

\subsubsection{DRL parameters} \label{RL-parameters}

For DQN we used the same hyperparameters described by \citet{alansary_automatic_2018}, except for the CNN architecture and the state sub-volume size, which were identical for both DQN and A3C. \autoref{tab:DRL_Parameters} lists the empirically chosen parameters for A3C. Sub-volumes of size (62,168,168) mm were selected to cover most structures in the reformatted view. The maximum values of rotation (5$^\circ$) and translation (5 mm) provided smooth transitions to the final reformatted planes, with a 100-step limit to ensure convergence. For network training, $\lambda$ was set to 0.025 to assign greater error magnitudes to the distance compared to the angles, thereby encouraging the agent to prioritize adjusting the location of the plane before refining its orientation. The remaining parameters used the default A3C values, as changing them did not significantly impact performance.

\begin{table}[h!]
\centering
\caption{Reinforcement learning and A3C hyperparameters}
\begin{tabular}{@{}lcl@{}}
\hline
\textbf{Name} & \textbf{Value} & \textbf{Description} \\
\hline
(D,H,W) & (62,168,168) & State sub-volumes dimensions in mm \\
$\omega_{\text{max}}$ & 5 & Maximum rotation in degrees \\
$d_{\text{max}}$ & 5 & Maximum translation in mm \\
$t_{\text{max}}$ & 100 & Maximum number of steps \\
$\lambda$ & 0.025 & Cost function weight \\
$\eta$ & 0.01 & Loss function entropy weight\\
$k_A$ & 6 & Advantage steps \\
\hline
\end{tabular}
\label{tab:DRL_Parameters}
\end{table}

\subsubsection{Network training}
Neural network operations were accelerated using a GPU. For the A3C implementation, we trained multiple CPU workers, each training a single agent and updating the parameters of a shared model. This setup enabled the network to learn policies across multiple volumes, reducing both training time and overfitting.

For each model (one per plane and cross-validation iteration), the A3C framework was trained using 10 AMD EPYC 7443 CPU cores and one NVIDIA RTX A6000 GPU, whereas the DQN implementation was trained using a single NVIDIA RTX A6000 GPU. 
Both algorithms employed the Adam optimizer \citep{kingma_adam_2017} with a constant learning rate of $10^{-5}$.The A3C framework was implemented in PyTorch, adapted from a public repository \citep{kostrikov_pytorch_2018}, while the DQN implementation of \citet{alansary_automatic_2018} was adapted to PyTorch and trained using the same hyperparameters.

\subsection{Plane reformatting evaluation}
\subsubsection{Plane reformatting performance}
For validation and testing, each algorithm was evaluated considering a 100-step episode. For testing, the selected model was the one with the lowest average value of $C(100)$ during validation, i.e., the validation model with the best performance in terms of angular and distance errors.

We measured the Euclidean distance between $P_{100}$ and $P_T$, and the angle between the normal vectors $\vec{n}_{100}$ and $\vec{n}_T$. The same metrics were evaluated between the annotations of both observers to measure inter-observer variability. Results are reported as mean $\pm$ standard deviation. 

We performed a statistical analysis to determine whether the angular and distance errors of AdaPR, relative to the observers, were comparable to the inter-observer errors. We assessed normality of the residuals using the Kolmogorov-Smirnov test \cite{massey1951kolmogorov}. Since the residuals did not meet the normality assumption, we applied a two-way Aligned Rank Transform (ART) ANOVA \cite{wobbrock2011aligned} to examine the effects of vessel (AAo, PA, RPA, LPA) and comparison type (AdaPR vs O1, AdaPR vs O2, Inter-Observer) on angular and distance errors. The ART procedure is a nonparametric approach appropriate for factorial designs when data violate normality assumptions. Post-hoc pairwise comparisons were performed using the Aligned Rank Transform Contrasts (ART-C) procedure \cite{elkin2021aligned} and p-values were adjusted using the Tukey method.

\subsubsection{Sensitivity analysis to state initialization}

For each algorithm and dataset, we performed 10 random initializations of the first state ($s_0$). These initializations matched those used during training (section \ref{sec:states}), consisting of random locations within 10\% of the volume size around the center, with random orientations. We then analyzed the performance across initializations.

\subsubsection{Sensitivity analysis to volume position and orientation}

Each algorithm was evaluated on datasets with different combinations of rigid transformations. Specifically, we applied rotations of 5°, 15°, and 25°, and translations of 5, 15, and 25 mm to all axes. We then computed the average performance of each algorithm across the augmented datasets.

\subsubsection{Single initialization and statistical analysis}

For reproducibility, we evaluated all datasets in one position, the center of the volume ($\vec{P}_0 = \{0,0,0\}$), and sagittal orientation ($\vec{n}_0 = \{1,0,0\}$). The predicted planes from this final evaluation were used for the statistical analysis and subsequent flow analysis.

\subsection{Flow analysis}

To segment vessels, we used a semi-automatic tool called Chan-Vese active contours \cite{chan_active_2001}, as in previous works \citep{montalba_variability_2018}. For each plane, the active contour was initialized with a disk centered at either the predicted vessel center ($P_{100}$) or the manual annotation ($P_T$). Each segmentation was visually inspected to ensure that it followed the correct vessel boundaries. A few scans showed segmentation errors due to the proximity of adjacent vessel walls, which required manual corrections.

To investigate how the segmentation differences between reformatting methods (AdaPR and observers) could affect flow evaluation, we computed the percentage differences in segmented areas and the Dice score \citep{taha2015metrics}. Since the segmentations corresponded to different planes, we first aligned the centers of mass of each segmented area and then computed the Dice score.


Using the heart rate recorded for each acquisition, we computed the flow in L/min for each segmented vessel. A regression analysis was used to quantify the coefficient of determination ($R^2$) for flow obtained from automated and manually placed planes, as well as to evaluate inter-observer variability. In addition, Bland-Altman plots were used to assess biases and establish limits of agreement.

To evaluate normality of the residuals, we used the Kolmogorov-Smirnov test \cite{massey1951kolmogorov}, and to assess homogeneity of variance, we applied Levene's test \cite{levene1960robust}. Since the residuals met the assumptions of normality and homogeneity of variance, we performed a two-way ANOVA to examine the effects of vessel (AAo, PA, RPA, LPA) and reformatting method (AdaPR, O1, O2) on flow measurements. Post-hoc pairwise comparisons were conducted using Tukey's HSD (Honest Significant Difference) test to identify specific differences between groups.


\subsection{Analysis software}

Bland-Altman plots were performed using GraphPad Prism v.9 (GraphPad Software, La Jolla, CA). Statistical analyzes were performed in R (R Development Core Team; \url{https://www.R-project.org}).

\section{Results}

\subsection{Plane Reformatting Performance}

\subsubsection{Comparison against alternative DRL methods}
When evaluated from a single initialization in the center of the volume, the proposed A3C-AdaPR framework achieved an average angular error of 6.32 $\pm$ 4.15$\degree$ and a distance error of 3.40 $\pm$ 2.75 mm relative to O1, outperforming DQN-VanillaPR (angular: 11.3 $\pm$ 8.44\degree, distance: 6.29 $\pm$ 6.01 mm) (\autoref{tab:comparison_reformatting_metrics}). Compared to A3C-VanillaPR, the improvements were smaller but still significant in angular error for most planes. 

\begin{table}[h!]
  \centering
  \caption{Plane reformatting errors relative to manual planes defined by observer 1, under two setups: (i) a single initialization and (ii) average across 10 random initializations per dataset. Bold values indicate the lowest error within each setup and metric.}
  \label{tab:comparison_reformatting_metrics}
  \sffamily
  \setlength{\tabcolsep}{4pt}
\newcolumntype{P}[1]{>{\raggedright\arraybackslash}p{#1}}
\begin{tabular}{@{} p{1.5cm} p{2.0cm} p{2.75cm} p{1.6cm} p{1.6cm} p{1.6cm} p{1.6cm} p{1.6cm} @{} }
    \toprule
    \multirow{2}{*}{Initialization} & \multirow{2}{*}{Algorithm} & \multirow{2}{*}{Metric} 
      & \multicolumn{5}{c}{Plane} \\
    \cmidrule(l){4-8}
      & & & \textit{AAo} & \textit{PA} & \textit{RPA} & \textit{LPA} & \textit{All} \\
    \midrule
    \multirow{6}{*}{\begin{tabular}[c]{@{}l@{}}Single \end{tabular}}
      & \multirow{2}{*}{A3C-AdaPR} 
        & Angular error (°) 
          & \textbf{6.13 ± 5.50} & \textbf{5.93 ± 3.21} & \textbf{6.25 ± 3.11} 
          & \textbf{6.97 ± 4.33} & \textbf{6.32 ± 4.15} \\
      &  & Distance error (mm) 
          & 5.15 ± 4.03 & 3.35 ± 1.90
          & \textbf{2.34 ± 1.44} & \textbf{2.76 ± 1.98} & \textbf{3.40 ± 2.75} \\
    \addlinespace
      & \multirow{2}{*}{DQN-VanillaPR} 
        & Angular error (°) 
          & 9.93 ± 6.51 & 13.1 ± 10.2 & 9.98 ± 5.73 
          & 12.1 ± 10.1 & 11.3 ± 8.44 \\
      &  & Distance error (mm) 
          & 4.66 ± 4.24 & 8.17 ± 6.88 
          & 5.07 ± 4.46 & 7.30 ± 7.20 & 6.29 ± 6.01 \\
    \addlinespace
      & \multirow{2}{*}{A3C-VanillaPR} 
        & Angular error (°) 
          & 7.42 ± 5.26 & 7.02 ± 4.19 & 8.98 ± 6.33 
          & 8.31 ± 5.22 & 7.93 ± 5.34 \\
      &  & Distance error (mm) 
          & \textbf{3.49 ± 3.90} & \textbf{2.73 ± 2.51} 
          & 4.16 ± 5.03 & 3.78 ± 3.73 & 3.54 ± 3.92 \\
    \midrule
    \multirow{6}{*}{\begin{tabular}[c]{@{}l@{}}Multiple\\ (10)\end{tabular}}
      & \multirow{2}{*}{A3C-AdaPR} 
        & Angular error (°) 
          & \textbf{6.36 ± 6.27} & \textbf{6.85 ± 4.73} & \textbf{6.27 ± 3.38} 
          & \textbf{7.78 ± 6.87} & \textbf{6.80 ± 5.51} \\
      &  & Distance error (mm)& 5.29 ± 4.48 & 3.48 ± 2.50 
          & \textbf{2.36 ± 1.33} & \textbf{3.10 ± 3.45} & \textbf{3.56 ± 3.33} \\
    \addlinespace
      & \multirow{2}{*}{DQN-VanillaPR} 
        & Angular error (°) 
          & 10.1 ± 7.75 & 16.3 ± 15.2 & 9.91 ± 6.30 
          & 12.4 ± 8.70 & 12.2 ± 10.4 \\
      &  & Distance error (mm)& \textbf{5.09 ± 5.59} & 11.2 ± 12.7 
          & 6.94 ± 11.3 & 8.02 ± 8.92 & 7.80 ± 10.2 \\
    \addlinespace
      & \multirow{2}{*}{A3C-VanillaPR} 
        & Angular error (°) 
          & 7.57 ± 5.37 & 7.24 ± 4.66 & 9.11 ± 6.58 
          & 8.54 ± 5.59 & 8.11 ± 5.64 \\
      &  & Distance error (mm)& 3.59 ± 3.96 & \textbf{2.88 ± 2.72} 
          & 4.15 ± 5.07 & 3.97 ± 4.10 & 3.65 ± 4.08 \\
    \bottomrule
  \end{tabular}
\end{table}
\begin{figure*}[!htb]
    \centering
    \includegraphics[width=0.9\textwidth,]{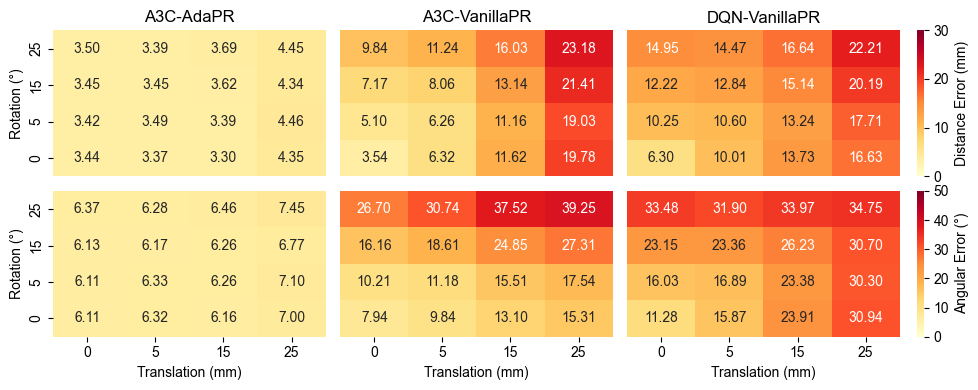}
    \caption{\textbf{Sensitivity of plane-reformatting accuracy to rigid transformations of the input volumes}. Heatmaps summarize average performance across all test volumes and planes when the same scans are perturbed by rotations of 5°, 15°, 25° and translations of 5, 15, 25 mm along all axes. Top panels report angular error (°); bottom panels report distance error (mm), both computed between the automated planes and the reference manual planes of observer 1. AdaPR maintains a near-constant error relative to the average error without rotations or translations, whereas VanillaPR variants show progressive degradation with larger perturbations.}
    \label{fig:RTexperiment}
\end{figure*}

\subsubsection{Sensitivity analysis to state initialization}

To assess robustness to different initializations, each algorithm was also evaluated in 10 random starting positions and orientations per dataset (\autoref{tab:comparison_reformatting_metrics}). Differences in average performance between single and multiple initializations were small for all algorithms, with minor increases in standard deviations. 


\subsubsection{Sensitivity analysis to volume position and orientation}

The second sensitivity experiment evaluated the performance under rigid transformations of the input volumes, including rotations of $5^\circ$, $15^\circ$, and $25^\circ$ and translations of 5, 15, and 25 mm along all axes (\autoref{fig:RTexperiment}). DQN-VanillaPR showed a clear degradation in both the angular and distance errors, even at the smallest perturbations, with errors increasing sharply at larger transformations. A3C-VanillaPR exhibited some robustness, but it was still degraded by stronger perturbations. In contrast, AdaPR maintained almost constant performance in all tested transformations, with average angular and distance errors remaining within 1.5$\degree$ and 1.1 mm with respect to the average error without rotations or translations.

\begin{table*}[!ht]
  \centering
  \caption{Metrics for reformatted planes: AdaPR vs observers and inter-observer errors.}
  \label{tab:metrics}
  \sffamily
  \begin{tabular}{@{} ll ccccc @{}}
    \toprule
    \multirow{2}{*}{Comparison} & \multirow{2}{*}{Metric}
      & \multicolumn{5}{c}{Plane} \\
    \cmidrule(l){3-7}
      &  & \textit{AAo} & \textit{PA} & \textit{RPA} & \textit{LPA} & \textit{All} \\
    \midrule
    \multirow{2}{*}{AdaPR vs Observer 1}
      & Angular error (°)
        & $6.13 \pm 5.50$ 
        & $5.93 \pm 3.21$
        & $6.25 \pm 3.11$
        & $6.97 \pm 4.33$
        & $6.32 \pm 4.15$ \\
      & Distance error (mm)
        & $5.15 \pm 4.03$
        & $3.35 \pm 1.90$
        & $2.34 \pm 1.44$
        & $2.76 \pm 1.98$
        & $3.40 \pm 2.75$ \\
    \addlinespace
    \multirow{2}{*}{AdaPR vs Observer 2}
      & Angular error (°)
        & $6.95 \pm 5.83$
        & $7.06 \pm 3.98$
        & $6.22 \pm 3.10$
        & $8.91 \pm 5.06$
        & $7.28 \pm 4.70$ \\
      & Distance error (mm)
        & $5.68 \pm 4.55$
        & $3.88 \pm 2.14$
        & $2.50 \pm 1.49$
        & $4.06 \pm 2.48$
        & $4.03 \pm 3.10$ \\
    \addlinespace
    \multirow{2}{*}{Inter‐observer}
      & Angular error (°)
        & $4.64 \pm 5.07$
        & $4.25 \pm 4.03$
        & $2.56 \pm 2.25$
        & $6.08 \pm 4.96$
        & $4.42 \pm 4.48$ \\
      & Distance error (mm)
        & $4.96 \pm 4.51$
        & $2.76 \pm 1.99$
        & $1.93 \pm 1.43$
        & $2.78 \pm 2.09$
        & $3.11 \pm 2.98$ \\
    \bottomrule
  \end{tabular}
\end{table*}

\subsubsection{Analysis of AdaPR plane reformatting}

For all datasets, planes defined by AdaPR compared to O1 yielded an angular error of $6.32 \pm 4.15^\circ$ and a distance error of $3.40 \pm 2.75$ mm (\autoref{tab:metrics}). Compared to O2, errors increased to $7.28 \pm 4.70^\circ$ and $4.03 \pm 3.10$ mm, respectively. Inter-observer variability showed angular and distance errors of $4.42 \pm 4.48^\circ$ and $3.11 \pm 2.98$ mm. Among all vessels, the PA plane demonstrated the lowest angular error ($5.93 \pm 3.21^\circ$), while LPA had the largest ($6.97 \pm 4.33^\circ$). For distance error, RPA showed the best performance ($2.34 \pm 1.44$ mm), whereas AAo exhibited the largest error ($5.15 \pm 4.03$ mm).

\begin{figure}[!ht]
    \centering
    \includegraphics[width=0.46\textwidth]{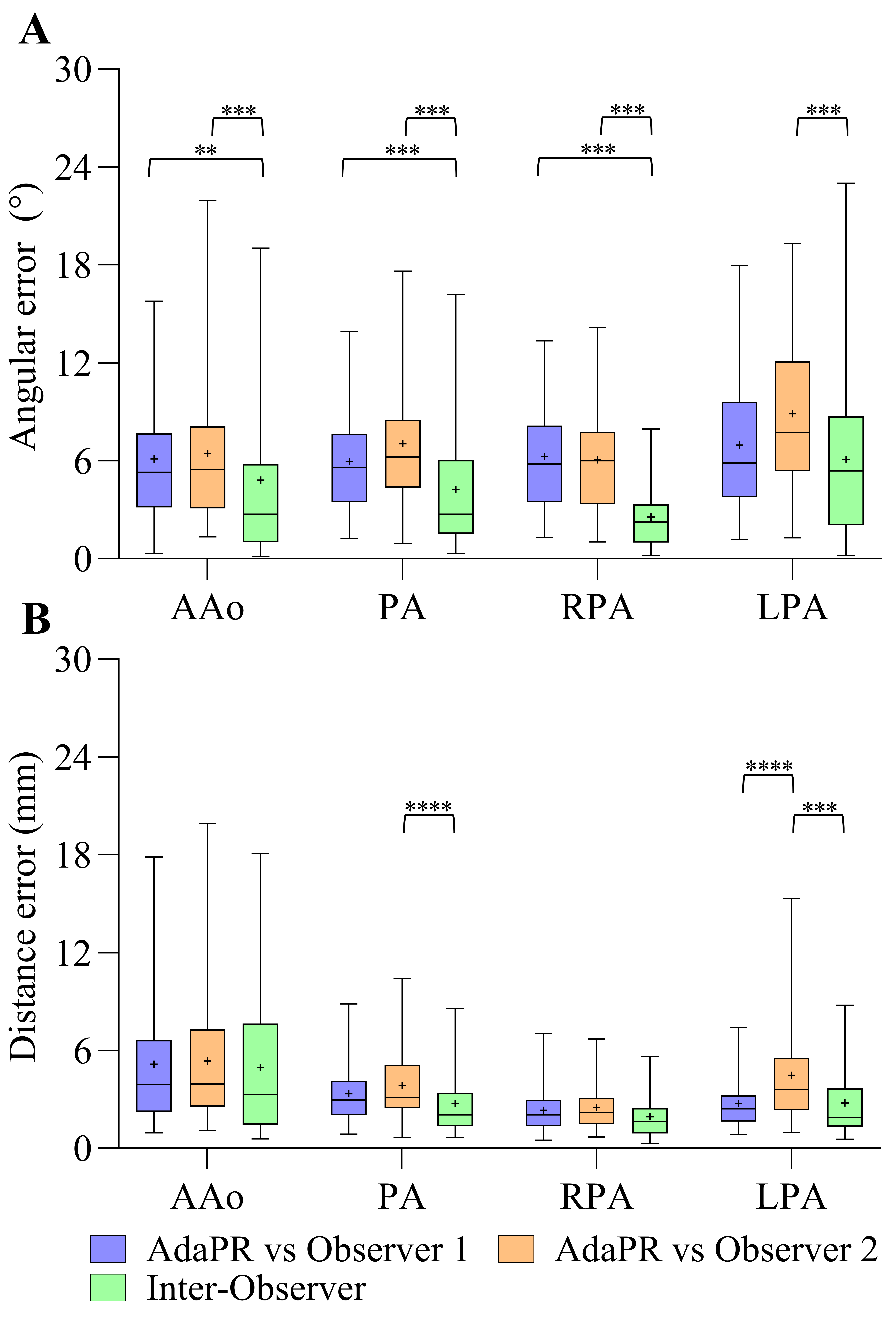}    
    \caption{\textbf{Distribution of plane reformatting metrics for AdaPR}.
    Whiskers range from the 2.5th to the 97.5th percentile. The line represents the median, and the cross represents the mean. \textbf{(A)} Angular error ($^\circ$). \textbf{(B)} Distance error (mm). **, ***, and  **** represent ART-C post-hoc contrasts with significance levels $<10^{-2}$,$<10^{-3}$, and $<10^{-4}$, respectively.}
    \label{fig:boxplots}
\end{figure}

For angular errors, the ART ANOVA revealed a significant interaction effect between comparison type (AdaPR vs O1, AdaPR vs O2, and Inter-Observer) and vessels ($p < 0.05$). Post-hoc ART-C pairwise comparisons with Tukey adjustment showed that for AAo, PA, and RPA vessels, both AdaPR angular errors relative to the observers were significantly larger than Inter-Observer angular errors ($p < 0.01$ for AAo; $p < 0.001$ for PA and RPA). Only for the LPA, AdaPR vs O1 did not show significant differences from Inter-Observer and AdaPR vs O2 angular errors ($p = 0.574$), whereas AdaPR vs O2 angular errors were significantly higher than Inter-Observer angular errors ($p < 0.001$). No significant differences were found between AdaPR vs O1 angular errors and AdaPR vs O2 angular errors ($p \geq 0.164$) (\autoref{fig:boxplots}).

For distance errors, the ART ANOVA revealed a significant interaction effect between comparison type and vessel ($p < 0.05$). Post-hoc ART-C tests identified three significant differences: for LPA, between Inter-observer and AdaPR vs O2 ($p < 0.001$), and between AdaPR vs O2 and AdaPR vs O1 ($p < 0.0001$); and for PA, between Inter-observer and AdaPR vs O2 ($p < 0.0001$). No significant differences were observed for AAo or RPA across comparison types ($p \geq 0.103$) (Figure~\ref{fig:boxplots}).

When stratified by population, healthy volunteers showed AdaPR errors of $6.00 \pm 4.17^\circ$ and $3.11 \pm 2.36$ mm relative to O1, compared with inter-observer errors of $4.51 \pm 4.25^\circ$ and $3.08 \pm 2.72$ mm (\autoref{tab:metrics_vol_and_patients}). In patients with congenital heart disease (bicuspid aortic valve, tetralogy of Fallot, or aortic coarctation), AdaPR errors relative to O1 were larger than in healthy volunteers ($7.31 \pm 3.98^\circ$ and $4.36 \pm 3.58$ mm), whereas inter-observer errors were similar to those of healthy volunteers ($4.14 \pm 5.14^\circ$ and $3.21 \pm 3.70$ mm).

\begin{table*}[!hb]
  \centering
  \caption{Metrics for reformatted planes: Comparison for AdaPR vs observer 1 and inter-observer errors grouped by patients and volunteers.}
  \label{tab:metrics_vol_and_patients}
  \sffamily
  \begin{tabular}{@{} ll ccccc @{}}
    \toprule
    \multirow{2}{*}{Comparison} & \multirow{2}{*}{Metric}
      & \multicolumn{5}{c}{Plane} \\
    \cmidrule(l){3-7}
      &  & \textit{AAo} & \textit{PA} & \textit{RPA} & \textit{LPA} & \textit{All} \\
    \midrule
    \multirow{2}{*}{AdaPR vs Observer 1 volunteers}
      & Angular error (°)
        & $6.01 \pm 3.52$
        & $5.70 \pm 2.93$
        & $6.30 \pm 3.82$
        & $5.96 \pm 4.16$
        & $5.96 \pm 4.16$ \\
      & Distance error (mm)
        & $3.40 \pm 2.04$
        & $2.17 \pm 1.33$
        & $2.33 \pm 0.93$
        & $3.09 \pm 2.36$
        & $3.09 \pm 2.36$ \\
    \addlinespace
    \multirow{2}{*}{AdaPR vs Observer 1 patients}
      & Angular error (°)
        & $5.69 \pm 2.02$
        & $7.99 \pm 3.10$
        & $9.24 \pm 5.22$
        & $7.47 \pm 3.95$
        & $7.47 \pm 3.95$ \\
      & Distance error (mm)
        & $3.22 \pm 1.39$
        & $2.89 \pm 1.67$
        & $4.21 \pm 3.44$
        & $4.41 \pm 3.59$
        & $4.41 \pm 3.59$ \\
    \addlinespace
    \multirow{2}{*}{Inter-observer volunteers}
      & Angular error (°)
        & $4.65 \pm 4.17$
        & $2.81 \pm 2.32$
        & $6.01 \pm 4.44$
        & $4.49 \pm 4.25$
        & $4.49 \pm 4.25$ \\
      & Distance error (mm)
        & $2.78 \pm 2.11$
        & $2.13 \pm 1.53$
        & $2.71 \pm 1.90$
        & $3.07 \pm 2.72$
        & $3.07 \pm 2.72$ \\
    \addlinespace
    \multirow{2}{*}{Inter-observer patients}
      & Angular error (°)
        & $2.99 \pm 3.34$
        & $1.77 \pm 1.87$
        & $6.31 \pm 6.56$
        & $4.20 \pm 5.16$
        & $4.20 \pm 5.16$ \\
      & Distance error (mm)
        & $2.73 \pm 1.59$
        & $1.31 \pm 0.78$
        & $3.03 \pm 2.68$
        & $3.25 \pm 3.72$
        & $3.25 \pm 3.72$ \\
    \bottomrule
  \end{tabular}
\end{table*}

\begin{figure}[h!]
    \centering
    \includegraphics[width=0.45\textwidth]{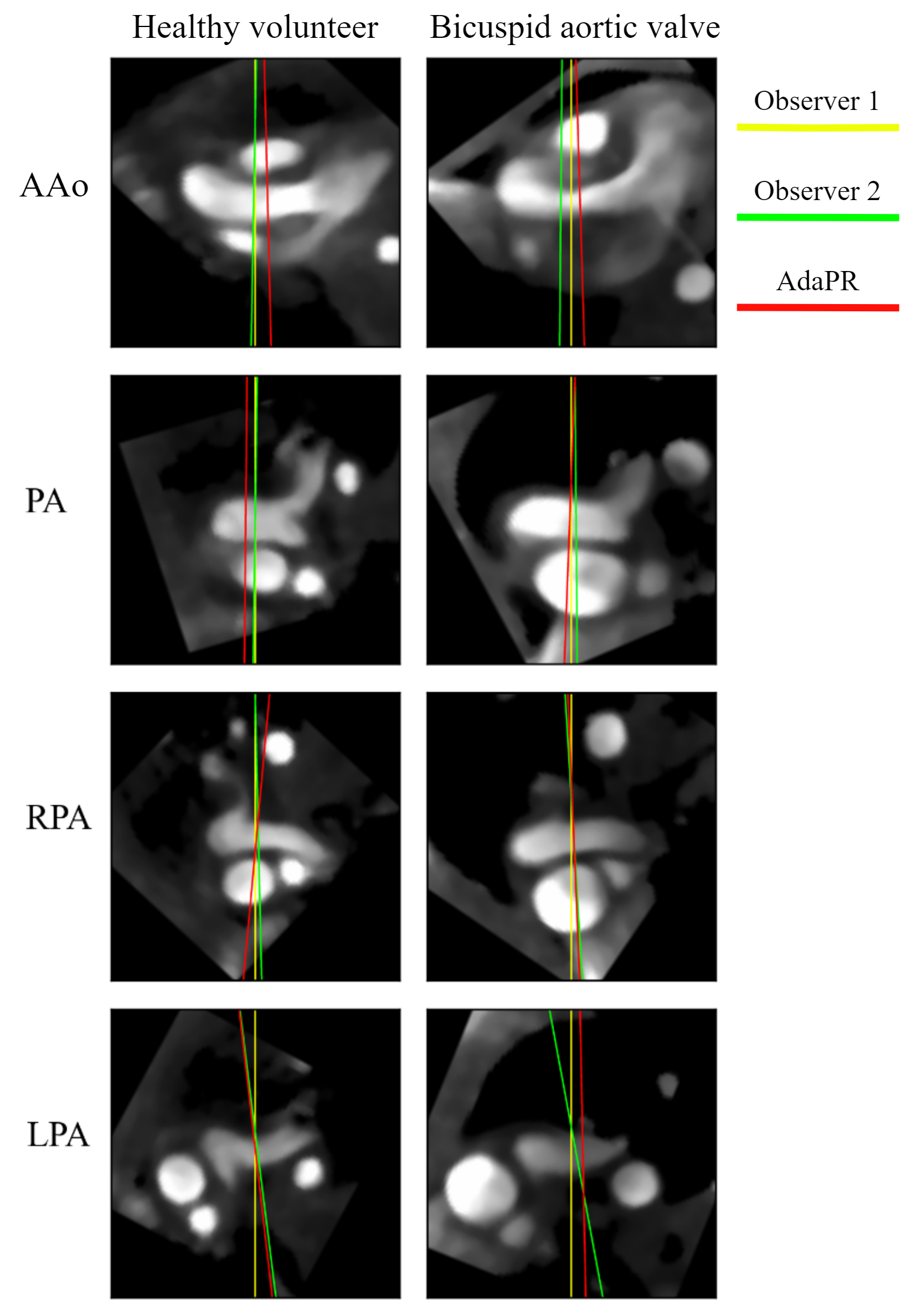}
    \caption{\textbf{Examples of plane reformatting}. Planes placed by observer 1 (O1), observer 2 (O2), and AdaPR planes are shown in yellow, green, and red, respectively. The left-hand column depicts a healthy volunteer; the right-hand column shows a patient with a bicuspid aortic valve (BAV).}
    \label{fig:vis-planes}
\end{figure}

\autoref{fig:vis-planes} shows examples of plane reformatting for a patient and a healthy volunteer showing great agreement between manual and automated planes.

\subsection{Flow Analysis}

\begin{table*}[h!]
\centering
\caption{Flow analysis. The first three rows show the average and standard deviation of flows measured by AdaPR and observers (O1 and O2). The subsequent three rows present the mean absolute percentage difference between the measurements. The final three rows show the coefficient of determination ($R^2$) from linear regression for each comparison. The last column summarizes the metrics across all planes.}
\label{tab:flow_analysis}
\begin{tabular}{lccccc}
\toprule
\textbf{Metric} & \textbf{AAo} & \textbf{PA} & \textbf{RPA} & \textbf{LPA} & \textbf{Average} \\
\midrule
Flow AdaPR [L/min] & $4.49 \pm 1.56$ & $4.55 \pm 1.49$ & $2.22 \pm 0.85$ & $1.98 \pm 0.72$ & $3.31 \pm 1.71$ \\
Flow O1 (L/min) & $4.48 \pm 1.54$ & $4.55 \pm 1.53$ & $2.23 \pm 0.86$ & $1.97 \pm 0.72$ & $3.31 \pm 1.72$ \\
Flow O2 (L/min) & $4.45 \pm 1.53$ & $4.55 \pm 1.51$ & $2.22 \pm 0.86$ & $1.98 \pm 0.71$ & $3.31 \pm 1.71$ \\
\midrule
Mean Diff. [\%] (AdaPR vs O1) & $4.07 \pm 3.26$ & $3.41 \pm 2.65$ & $5.80 \pm 4.66$ & $5.72 \pm 5.42$ & $4.75 \pm 4.25$ \\
Mean \% Diff. (AdaPR vs O2) & $4.29 \pm 3.62$ & $3.51 \pm 2.64$ & $6.46 \pm 5.11$ & $6.48 \pm 7.49$ & $5.18 \pm 5.20$ \\
Mean \% Diff. (O1 vs O2) & $4.58 \pm 4.57$ & $2.96 \pm 2.75$ & $5.81 \pm 5.01$ & $6.56 \pm 7.09$ & $4.97 \pm 5.25$ \\
\midrule
$R^2$ (AdaPR vs O1) & $0.981$ & $0.986$ & $0.957$ & $0.963$ & $0.972$ \\
$R^2$ (AdaPR vs O2) & $0.982$ & $0.985$ & $0.956$ & $0.947$ & $0.968$ \\
$R^2$ (O1 vs O2) & $0.977$ & $0.986$ & $0.966$ & $0.948$ & $0.969$ \\
\bottomrule
\end{tabular}
\end{table*}

\begin{figure}[!h]
    \centering
    \includegraphics[width=0.5\textwidth,]{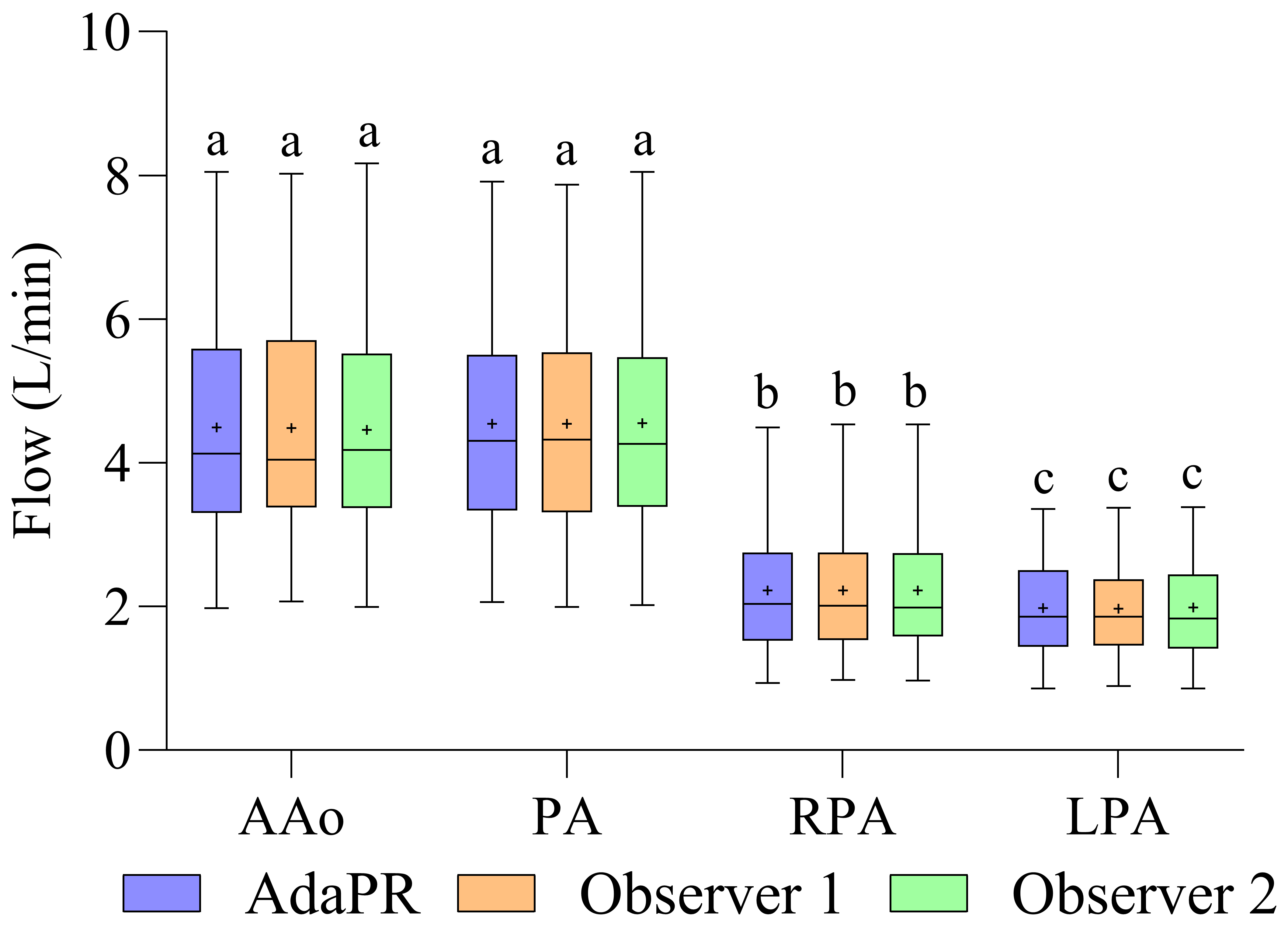}
    \caption{Flow measurements (L/min) at four vessels (AAo: ascending aorta; PA: pulmonary artery; RPA: right pulmonary artery; LPA: left pulmonary artery) for AdaPR, Observer 1, and Observer 2. Box plots show median (line), mean (+), interquartile range (box),and whiskers range from the 2.5th to the 97.5th percentile. Different letters represent groups; box plots with the same letter have no significant differences based on two-way ANOVA post-hoc comparisons (p $>$ 0.05).}
    \label{fig:Flow-Boxplots}
\end{figure}

Flow measurements obtained with AdaPR (\autoref{tab:flow_analysis}) showed strong agreement with manual quantification. The average percentage differences between AdaPR and O1 was $4.748 \pm 4.254 $ and between AdaPR and O2 was $5.181 \pm 5.196$, values that are comparable to the inter-observer difference ($4.971 \pm 5.246$). AdaPR vs O1 achieved an overall coefficient of determination of $R^2=0.972$, closely matching both the inter-observer agreement ($R^2=0.969$) and the AdaPR vs O2 comparison ($R^2=0.968$). Across individual planes, $R^2$ values ranged from 0.947 to 0.986 (\autoref{tab:flow_analysis}), indicating consistently high concordance between automated and manual measurements.

Two-way ANOVA revealed no significant differences between reformatting methods (AdaPR, O1, and O2) ($p=0.663$), nor a significant interaction between reformatting methods and planes ($p=0.809$). Flow differences were only observed between planes, with RPA and LPA showing different flow distributions (\autoref{fig:Flow-Boxplots}).

The segmentation variability analysis showed mean absolute area differences of 8.6\% for AdaPR vs O1 (median 6.7\%), 9.4\% for AdaPR vs O2 (median 7.0\%), and 8.2\% for O1 vs O2 (median 5.7\%). Dice coefficients after center-of-mass alignment were 0.912 (AdaPR vs O1), 0.909 (AdaPR vs O2), and 0.918 (O1 vs O2). These results indicate that AdaPR yields segmentation consistency comparable to inter-observer variability.

\begin{figure*}[!h]
    \centering
    \includegraphics[width=0.999\textwidth,]{Bland_Altman_Analysis.png}
    \caption{Bland–Altman plots showing percentage flow differences for each vessel (AAo, PA, RPA, LPA). The first row compares AdaPR vs Observer 1, the second row AdaPR vs Observer 2, and the third row Observer 1 vs Observer 2. The solid blue line indicates the mean bias, and the dashed red lines denote the 95\% limits of agreement. Across vessels, AdaPR demonstrates minimal bias and agreement comparable to inter-observer variability.}
    \label{fig:BA-flows}
\end{figure*}

Bland–Altman analysis (\autoref{fig:BA-flows}) showed similar limits of agreement (LoAs) for percentage flow differences in both AdaPR versus observers and inter-observer comparisons: approximately ±10\% for AAo and PA, and ±15–20\% for RPA and LPA. Overall, these results demonstrate that AdaPR achieves agreement with manual flow quantification comparable to that observed between human observers, and that the wider LoAs in the pulmonary arteries reflect expected challenges in resolution when measuring flow in smaller vessels.

\subsection{Training and Evaluation Times}
On average, training A3C-VanillaPR, DQN-VanillaPR, and A3C-AdaPR for a single plane required 12, 15, and 24 hours, respectively.  Depending on the available resources, additional CPU cores or GPUs could accelerate training without compromising performance. Pre-processing of all volumes for a single scan required between 3 and 5 seconds, and evaluation required between 3 and 4 seconds per plane.

\section{Discussion}
\subsection{Plane reformatting}

Our study demonstrates excellent performance of AdaPR for plane reformatting in 4D flow MRI, particularly superior to the benchmark DRL framework, DQN-VanillaPR. This improved accuracy can be attributed to two main factors: the advantages of the Asynchronous Advantage Actor-Critic (A3C) algorithm and the implementation of a local (adaptive) coordinate system. A3C algorithm is an on-policy method that refines a single policy through active exploration, resulting in more stable and robust learning than the off-policy DQN approach. We initially trained a DQN-AdaPR variant, where the agent operated in the local coordinate framework using the DQN algorithm. However, the larger and more complex state space introduced by AdaPR prevented DQN from converging to a stable solution. Consequently, we adopted the A3C algorithm for AdaPR to ensure more stable and consistent training.

The adaptive local coordinate system further enhanced performance by allowing the agent to navigate within volumes of arbitrary position and orientation, thereby improving both accuracy and generalization.

A key finding was the robustness of all DRL algorithms to state initialization. When evaluated from 10 random starting positions and orientations per dataset, the average performance of all algorithms showed only minor deviations from that observed with a single initialization. This behavior was expected, as the models were trained with randomized initializations around the center of the volume, enabling them to generalize effectively to different starting configurations.

The results also highlight a critical distinction between frameworks when subjected to rotations and translations. The VanillaPR framework showed a clear degradation in plane reformatting accuracy (both angular and distance errors) as perturbations increased. In contrast, AdaPR maintained nearly constant performance, with average angular and distance errors within 1.5$\degree$ and 1.1 mm of the unperturbed baseline, respectively. This greater generalizability underscores the potential of AdaPR for clinical applications, where imaging volumes may not be consistently aligned.

For distance error, AdaPR performed comparably to manual reformatting (\autoref{fig:boxplots}). Significant differences were only observed between AdaPR vs O2 and the inter-observer comparison for LPA and PA, whereas no significant differences were found for the remaining comparisons between AdaPR vs O1 or AdaPR vs O2 and the inter-observer errors. This suggests that the spatial precision of AdaPR typically falls within the range of human variability. In contrast, for angular errors, both AdaPR vs O1 and AdaPR vs O2 showed significantly larger discrepancies than the inter-observer comparison for most vessels, indicating room for improvement in orientation alignment. However, inter-observer differences reported in the literature vary considerably. For example, Corrado et al. \citep{corrado_automatic_2022} reported substantially larger inter-observer errors ($8.3 \pm 5.8$ mm for distance error and $13.9 \pm 7.8^\circ$ for angular error) than those found in our study. Establishing clinically validated benchmarks for inter-observer variability in plane reformatting would enable fairer and more standardized cross-study comparisons.



To compare AdaPR with other reformatting methods besides DRL, we assessed average performance metrics reported across relevant studies (\autoref{tab:comparison_methods}). We compared the average distance and angular errors for all reformatted planes in each study. Differences between AdaPR and the alternative methods were evaluated using the t-test based on the published means, standard deviations, and sample sizes. For studies using 4D flow MRI datasets, we also compared $R^2$ values of flow measurements in the ascending aorta.

\begin{table}[!ht]
\centering
\caption{Comparison with non-DRL methods. Error metrics correspond to the average values reported in each publication.}
\label{tab:comparison_methods}
\begin{tabular}{@{}llccc@{}}
  \toprule
  \textbf{Method} & \textbf{Data type} & \textbf{Distance error (mm)} 
    & \textbf{Angular error ($^\circ$)} & \textbf{Flow aorta ($R^2$)} \\
  \midrule
  AdaPR (Ours) & 4D flow MRI & $\mathbf{3.4\pm2.8}$ 
       & $\mathbf{6.3\pm4.2}$ & $\mathbf{0.98}$ \\
  3D CNN \citep{corrado_automatic_2022} 
       & 4D flow MRI & $9.0\pm7.3^\dagger$ 
       & $12.3\pm7.2^\dagger$ & $0.74$ \\
  Atlas registration \citep{bustamante_atlas-based_2015} 
       & 4D flow MRI & $\ast$ & $\ast$ & $0.93$ \\
  DL landmarks \citep{blansit_deep_2019} 
       & Cardiac Cine MRI & $7.5\pm8.2^\dagger$ 
       & $6.4\pm5.9$ & --- \\
  Atlas prompting \citep{fan2025avp}** 
       & Cardiac CT & $7.0\pm6.3^\dagger$ & $8.2\pm8.1^\dagger$ & --- \\
  \bottomrule
  \multicolumn{5}{p{.99\columnwidth}}{
  \small $\ast$ Values not reported.
  
  ** Atlas prompting errors using rigid registration
  
    $^\dagger$ Denotes significant difference ($p<0.05$) compared to AdaPR.}
\end{tabular}%
\end{table}

We found significant differences between AdaPR and all previous methods ($p<0.05$), except for DL landmark localization \citep{blansit_deep_2019} in terms of angular error ($p=0.84$), for which we observed a similar average angular error, but with lower variability.

\subsection{Flow measurements}

Our results showed no significant differences in flow quantification across reformatting methods (AdaPR, O1, and O2), as demonstrated by a two-way ANOVA on the full dataset (\autoref{fig:Flow-Boxplots}). The average percentage differences between AdaPR and observer measurements, as well as inter-observer differences, remained close to $\sim$5\%. Notably, AdaPR demonstrated superior agreement with manual aortic flow quantification ($R^2 = 0.98$) compared with alternative methods such as the 3D CNN \cite{corrado_automatic_2022} ($R^2 = 0.74$) and atlas registration \cite{bustamante_atlas-based_2015} ($R^2 = 0.93$) (\autoref{tab:comparison_methods}).

Although angular errors between AdaPR and observers were generally higher than inter-observers angular differences, these discrepancies had a negligible impact on flow measurements. This is consistent with previous findings \citep{casciaro_4d_2021}, which showed that angulation changes exceeding $25^\circ$, or ROI diameter variation of $\pm10\%$ are needed to produce substantial effects on aortic flow measurements. However, these factors were analyzed individually, whereas in practice they may occur simultaneously, suggesting that smaller deviations should still be considered when targeting  high-accuracy flow assessment.

Bland–Altman analysis (\autoref{fig:BA-flows}) further showed that AdaPR achieves LoAs for percentage flow differences comparable to inter-observer variability: approximately ±10\% for AAo and PA, and ±15–20\% for RPA and LPA. The larger LoAs observed in RPA and LPA likely reflect spatial resolution constraints, as these vessel have smaller cross-sectional areas \cite{dyverfeldt_4d_2015}.

\subsection{Limitations}

Our framework relies on supervised training, meaning that reformatting accuracy depends on two key factors: the size and diversity of the training data, and the quality of the manual labels used for supervision. AdaPR was evaluated on a relatively small dataset of  88 scans, only 25\% of which were patients. Although errors were slightly larger in patients ($7.47^\circ$ and $4.41$ mm) (\autoref{tab:metrics_vol_and_patients} Appendix), this may reflect the limited representation of patients. Increasing the number of patient studies is likely to reduce this difference. Nevertheless, expanding clinical imaging datasets is inherently challenging due to strict privacy regulations and the labor-intensive nature of manual labeling. As dataset sizes increase, manual labeling becomes increasingly impractical, posing a persistent bottleneck in medical imaging analysis.

In addition, pre-processing steps such as contrast enhancement and interpolation to isotropic resolution are required when evaluating new datasets. Although these procedures aim to reduce the variability across datasets, they may discard valuable information that could further improve reformatting accuracy.

\subsection{Future directions}

To improve performance, future work could incorporate additional training objectives, such as maximizing through-plane flow, to reduce biases introduced by limited datasets or imperfect labels. Collaborative agent approaches \citep{huang_searching_2020} could also mitigate variability across planes and enhance overall plane reformatting accuracy. Further evaluations in patients with discordant or complex cardiac anatomy are also needed to confirm the generalizability of AdaPR.

A key advantage of our approach is its adaptability beyond 4D flow MRI. By learning from volumetric navigation rather than fixed landmarks, it can be applied to other MRI modalities or CT angiography, enabling broad clinical and research applications in automatic plane reformatting.

\section{Conclusion}

We developed AdaPR, a robust plane reformatting framework that is effective in volumes with different positions, orientations, scanner vendors, and type of patients. AdaPR was validated on 4D flow MRI, revealing no significant disparities in flow quantification compared to manual quantification. The reinforcement learning frameworks for plane reformatting presented in this study are open-source and available at \url{https://github.com/JavierBZ/Adaptive-MRI-Plane-Reformatting}

\section*{Declaration of competing interest}
The authors declare that they have no known competing financial interests or personal relationships that could have appeared to influence the work reported in this paper.

\section*{Data availability}

Due to privacy concerns, the authors do not have permission to share data.

\section*{Acknowledgments}
\noindent
J.B. was funded by the National Agency for Research and Development (ANID) / Scholarship Program / DOCTORADO BECAS CHILE/2022 – 21220454 and by ANID - Millennium Science Initiative Program - ICN2021\_004. J.S. thanks the Department of Medical Imaging and Radiation Sciences at Monash University. C.T thanks Fondecyt 1231535. C.T. and J.S thank ANID - Millennium Science Initiative Program - ICN2021\_004. P.I. thanks Fondecyt 1210747.

\section*{Ethics statement}
4D flow MRI data were retrospectively obtained from institutional review board (IRB)–approved studies at four institutions. All studies were conducted in accordance with the Declaration of Helsinki. Study numbers, institution, and approval dates were as follows: REB13-090 (University of Calgary, Calgary, Canada; approval date: 14 August 2014), 20210122175843 (Hôpital Necker Enfants Malades, Paris, France; approval date: 21 January 2021), PR(AG)363/2016 (Hospital Universitari Vall d’Hebron, Barcelona, Spain; approval date: 16 December 2016), and 171127001 (Pontificia Universidad Católica de Chile, Santiago, Chile; approval date: 20 June 2019). Written informed consent was obtained from all participants, except for study 20210122175843 (Hôpital Necker Enfants Malades, Paris, France), for which, in accordance with European regulations on the use of fully anonymized data, the requirement for informed consent was waived.

\printcredits

\section*{Declaration of generative AI and AI-assisted technologies in the manuscript preparation process}

During the preparation of this work the authors used ChatGPT (OpenAI, San Francisco, CA, USA) in order to improve language and readability. After using this tool/service, the authors reviewed and edited the content as needed and take full responsibility for the content of the published article.
\bibliographystyle{model1-num-names}
\bibliography{cas-refs}

@article{grondman_survey_2012,
	title = {A {Survey} of {Actor}-{Critic} {Reinforcement} {Learning}: {Standard} and {Natural} {Policy} {Gradients}},
	volume = {42},
	issn = {1094-6977, 1558-2442},
	shorttitle = {A {Survey} of {Actor}-{Critic} {Reinforcement} {Learning}},
	url = {http://ieeexplore.ieee.org/document/6392457/},
	doi = {10.1109/TSMCC.2012.2218595},
	language = {en},
	number = {6},
	urldate = {2023-08-09},
	journal = {IEEE Transactions on Systems, Man, and Cybernetics, Part C (Applications and Reviews)},
	author = {Grondman, Ivo and Busoniu, Lucian and Lopes, Gabriel A. D. and Babuska, Robert},
	year = {2012},
	pages = {1291--1307},
}

@article{garcia_aortic_2021,
	title = {Aortic and mitral flow quantification using dynamic valve tracking and machine learning: {Prospective} study assessing static and dynamic plane repeatability, variability and agreement},
	volume = {10},
	issn = {2048-0040, 2048-0040},
	shorttitle = {Aortic and mitral flow quantification using dynamic valve tracking and machine learning},
	url = {http://journals.sagepub.com/doi/10.1177/2048004021999900},
	doi = {10.1177/2048004021999900},
	language = {en},
	journal = {JRSM Cardiovascular Disease},
	author = {Garcia, Julio and Beckie, Kailey and Hassanabad, Ali F and Sojoudi, Alireza and White, James A},
	year = {2021},
	pages = {204800402199990},
}

@article{gu2005pc,
  title={PC VIPR: a high-speed 3D phase-contrast method for flow quantification and high-resolution angiography},
  author={Gu, Tianliang and Korosec, Frank R and Block, Walter F and Fain, Sean B and Turk, Quill and Lum, Darren and Zhou, Yong and Grist, Thomas M and Haughton, Victor and Mistretta, Charles A},
  journal={American journal of neuroradiology},
  volume={26},
  number={4},
  pages={743--749},
  year={2005},
  publisher={American Journal of Neuroradiology}
}

@article{sotelo_fully_2022,
	title = {Fully {Three}-{Dimensional} {Hemodynamic} {Characterization} of {Altered} {Blood} {Flow} in {Bicuspid} {Aortic} {Valve} {Patients} {With} {Respect} to {Aortic} {Dilatation}: {A} {Finite} {Element} {Approach}},
	volume = {9},
	issn = {2297-055X},
	shorttitle = {Fully {Three}-{Dimensional} {Hemodynamic} {Characterization} of {Altered} {Blood} {Flow} in {Bicuspid} {Aortic} {Valve} {Patients} {With} {Respect} to {Aortic} {Dilatation}},
	url = {https://www.frontiersin.org/articles/10.3389/fcvm.2022.885338/full},
	doi = {10.3389/fcvm.2022.885338},
	language = {en},
	journal = {Frontiers in Cardiovascular Medicine},
	author = {Sotelo, Julio and Franco, Pamela and Guala, Andrea and Dux-Santoy, Lydia and Ruiz-Muñoz, Aroa and Evangelista, Arturo and Mella, Hernan and Mura, Joaquín and Hurtado, Daniel E. and Rodríguez-Palomares, José F. and Uribe, Sergio},
	year = {2022},
	pages = {885338},
}

@article{montalba_variability_2018,
	title = {Variability of {4D} flow parameters when subjected to changes in {MRI} acquisition parameters using a realistic thoracic aortic phantom},
	volume = {79},
	issn = {0740-3194, 1522-2594},
	url = {https://onlinelibrary.wiley.com/doi/10.1002/mrm.26834},
	doi = {10.1002/mrm.26834},
	language = {en},
	number = {4},
	journal = {Magnetic Resonance in Medicine},
	author = {Montalba, Cristian and Urbina, Jesus and Sotelo, Julio and Andia, Marcelo E. and Tejos, Cristian and Irarrazaval, Pablo and Hurtado, Daniel E. and Valverde, Israel and Uribe, Sergio},
	year = {2018},
	pages = {1882--1892},
}

@article{chan_active_2001,
	title = {Active contours without edges},
	volume = {10},
	issn = {10577149},
	url = {http://ieeexplore.ieee.org/document/902291/},
	doi = {10.1109/83.902291},
	language = {en},
	number = {2},
	journal = {IEEE Transactions on Image Processing},
	author = {Chan, T.F. and Vese, L.A.},
	year = {2001},
	pages = {266--277},
}

@misc{kostrikov_pytorch_2018,
	title = {{PyTorch} {Implementations} of {Asynchronous} {Advantage} {Actor} {Critic}},
	url = {https://github.com/ikostrikov/pytorch-a3c},
	publisher = {GitHub},
	author = {Kostrikov, Ilya},
	year = {2018},
	note = {GitHub repository},
}

@misc{kingma_adam_2017,
	title = {Adam: {A} {Method} for {Stochastic} {Optimization}},
	shorttitle = {Adam},
	url = {http://arxiv.org/abs/1412.6980},
	language = {en},
	publisher = {arXiv},
	author = {Kingma, Diederik P. and Ba, Jimmy},
	year = {2017},
	note = {Published as a conference paper at the International Conference for Learning Representations, 2015},
}

@article{hochreiter_long_1997,
	title = {Long {Short}-{Term} {Memory}},
	volume = {9},
	issn = {0899-7667, 1530-888X},
	url = {https://direct.mit.edu/neco/article/9/8/1735-1780/6109},
	doi = {10.1162/neco.1997.9.8.1735},
	language = {en},
	number = {8},
	journal = {Neural Computation},
	author = {Hochreiter, Sepp and Schmidhuber, Jürgen},
	year = {1997},
	pages = {1735--1780},
}

@inproceedings{he_deep_2016,
	address = {Las Vegas, NV, USA},
	title = {Deep {Residual} {Learning} for {Image} {Recognition}},
	isbn = {978-1-4673-8851-1},
	url = {http://ieeexplore.ieee.org/document/7780459/},
	doi = {10.1109/CVPR.2016.90},
	language = {en},
	booktitle = {2016 {IEEE} {Conference} on {Computer} {Vision} and {Pattern} {Recognition} ({CVPR})},
	publisher = {IEEE},
	author = {He, Kaiming and Zhang, Xiangyu and Ren, Shaoqing and Sun, Jian},
	year = {2016},
	pages = {770--778},
}

@article{bissell_4d_2023,
	title = {{4D} {Flow} cardiovascular magnetic resonance consensus statement: 2023 update},
	volume = {25},
	issn = {1532-429X},
	shorttitle = {{4D} {Flow} cardiovascular magnetic resonance consensus statement},
	url = {https://jcmr-online.biomedcentral.com/articles/10.1186/s12968-023-00942-z},
	doi = {10.1186/s12968-023-00942-z},
	language = {en},
	number = {1},
	journal = {Journal of Cardiovascular Magnetic Resonance},
	author = {Bissell, Malenka M. and Raimondi, Francesca and Ait Ali, Lamia and Allen, Bradley D. and Barker, Alex J. and Bolger, Ann and Burris, Nicholas and Carhäll, Carl-Johan and Collins, Jeremy D. and Ebbers, Tino and Francois, Christopher J. and Frydrychowicz, Alex and Garg, Pankaj and Geiger, Julia and Ha, Hojin and Hennemuth, Anja and Hope, Michael D. and Hsiao, Albert and Johnson, Kevin and Kozerke, Sebastian and Ma, Liliana E. and Markl, Michael and Martins, Duarte and Messina, Marci and Oechtering, Thekla H. and Van Ooij, Pim and Rigsby, Cynthia and Rodriguez-Palomares, Jose and Roest, Arno A. W. and Roldán-Alzate, Alejandro and Schnell, Susanne and Sotelo, Julio and Stuber, Matthias and Syed, Ali B. and Töger, Johannes and Van Der Geest, Rob and Westenberg, Jos and Zhong, Liang and Zhong, Yumin and Wieben, Oliver and Dyverfeldt, Petter},
	year = {2023},
	pages = {40},
}

@article{dyverfeldt_4d_2015,
	title = {{4D} flow cardiovascular magnetic resonance consensus statement},
	volume = {17},
	issn = {1532-429X},
	url = {https://jcmr-online.biomedcentral.com/articles/10.1186/s12968-015-0174-5},
	doi = {10.1186/s12968-015-0174-5},
	language = {en},
	number = {1},
	journal = {Journal of Cardiovascular Magnetic Resonance},
	author = {Dyverfeldt, Petter and Bissell, Malenka and Barker, Alex J. and Bolger, Ann F. and Carlhäll, Carl-Johan and Ebbers, Tino and Francois, Christopher J. and Frydrychowicz, Alex and Geiger, Julia and Giese, Daniel and Hope, Michael D. and Kilner, Philip J. and Kozerke, Sebastian and Myerson, Saul and Neubauer, Stefan and Wieben, Oliver and Markl, Michael},
	year = {2015},
	pages = {72},
}

@misc{mnih_asynchronous_2016,
	title = {Asynchronous {Methods} for {Deep} {Reinforcement} {Learning}},
	url = {http://arxiv.org/abs/1602.01783},
	language = {en},
	publisher = {arXiv},
	author = {Mnih, Volodymyr and Badia, Adrià Puigdomènech and Mirza, Mehdi and Graves, Alex and Lillicrap, Timothy P. and Harley, Tim and Silver, David and Kavukcuoglu, Koray},
	year = {2016},
	note = {arXiv:1602.01783 [cs]},
}

@misc{huang_searching_2020,
	title = {Searching {Collaborative} {Agents} for {Multi}-plane {Localization} in {3D} {Ultrasound}},
	url = {http://arxiv.org/abs/2007.15273},
	language = {en},
	publisher = {arXiv},
	author = {Huang, Yuhao and Yang, Xin and Li, Rui and Qian, Jikuan and Huang, Xiaoqiong and Shi, Wenlong and Dou, Haoran and Chen, Chaoyu and Zhang, Yuanji and Luo, Huanjia and Frangi, Alejandro and Xiong, Yi and Ni, Dong},
	year = {2020},
	note = {Accepted by MICCAI 2020},
}

@misc{dou_agent_2019,
	title = {Agent with {Warm} {Start} and {Active} {Termination} for {Plane} {Localization} in {3D} {Ultrasound}},
	url = {http://arxiv.org/abs/1910.04331},
	language = {en},
	publisher = {arXiv},
	author = {Dou, Haoran and Yang, Xin and Qian, Jikuan and Xue, Wufeng and Qin, Hao and Wang, Xu and Yu, Lequan and Wang, Shujun and Xiong, Yi and Heng, Pheng-Ann and Ni, Dong},
	year = {2019},
	note = {Accepted by MICCAI 2019},
}

@misc{alansary_automatic_2018,
	title = {Automatic {View} {Planning} with {Multi}-scale {Deep} {Reinforcement} {Learning} {Agents}},
	url = {http://arxiv.org/abs/1806.03228},
	language = {en},
	publisher = {arXiv},
	author = {Alansary, Amir and Folgoc, Loic Le and Vaillant, Ghislain and Oktay, Ozan and Li, Yuanwei and Bai, Wenjia and Passerat-Palmbach, Jonathan and Guerrero, Ricardo and Kamnitsas, Konstantinos and Hou, Benjamin and McDonagh, Steven and Glocker, Ben and Kainz, Bernhard and Rueckert, Daniel},
	year = {2018},
	note = {Accepted by MICCAI 2018},
}

@article{blansit_deep_2019,
	title = {Deep {Learning}–based {Prescription} of {Cardiac} {MRI} {Planes}},
	volume = {1},
	issn = {2638-6100},
	url = {http://pubs.rsna.org/doi/10.1148/ryai.2019180069},
	doi = {10.1148/ryai.2019180069},
	language = {en},
	number = {6},
	journal = {Radiology: Artificial Intelligence},
	author = {Blansit, Kevin and Retson, Tara and Masutani, Evan and Bahrami, Naeim and Hsiao, Albert},
	year = {2019},
	pages = {e180069},
}

@article{corrado_automatic_2022,
	title = {Automatic measurement plane placement for {4D} {Flow} {MRI} of the great vessels using deep learning},
	volume = {17},
	issn = {1861-6410, 1861-6429},
	url = {https://link.springer.com/10.1007/s11548-021-02475-1},
	doi = {10.1007/s11548-021-02475-1},
	language = {en},
	number = {1},
	journal = {International Journal of Computer Assisted Radiology and Surgery},
	author = {Corrado, Philip A. and Seiter, Daniel P. and Wieben, Oliver},
	year = {2022},
	pages = {199--210},
}

@article{pizer_adaptive_1987,
	title = {Adaptive histogram equalization and its variations},
	volume = {39},
	issn = {0734-189X},
	number = {3},
	journal = {Computer Vision, Graphics, and Image Processing},
	author = {Pizer, Stephen M and Amburn, E Philip and Austin, John D and Cromartie, Robert and Geselowitz, Ari and Greer, Trey and ter Haar Romeny, Bart and Zimmerman, John B and Zuiderveld, Karel},
	year = {1987},
	note = {Publisher: Elsevier},
	pages = {355--368},
}

@article{bustamante_atlas-based_2015,
  title={Atlas-based analysis of 4D flow CMR: automated vessel segmentation and flow quantification},
  author={Bustamante, Mariana and Petersson, Sven and Eriksson, Jonatan and Alehagen, Urban and Dyverfeldt, Petter and Carlh{\"a}ll, Carl-Johan and Ebbers, Tino},
  journal={Journal of Cardiovascular Magnetic Resonance},
  volume={17},
  number={1},
  pages={87},
  year={2015},
  publisher={Elsevier}
}

@article{uribe_four-dimensional_2009,
	title = {Four-dimensional ({4D}) flow of the whole heart and great vessels using real-time respiratory self-gating: {Real}-{Time} {Self}-{Gating} {4D} {Flow} {Imaging}},
	volume = {62},
	issn = {07403194},
	shorttitle = {Four-dimensional ({4D}) flow of the whole heart and great vessels using real-time respiratory self-gating},
	url = {https://onlinelibrary.wiley.com/doi/10.1002/mrm.22090},
	doi = {10.1002/mrm.22090},
	language = {en},
	number = {4},
	journal = {Magnetic Resonance in Medicine},
	author = {Uribe, Sergio and Beerbaum, Philipp and Sørensen, Thomas Sangild and Rasmusson, Allan and Razavi, Reza and Schaeffter, Tobias},
	year = {2009},
	pages = {984--992},
}

@article{casciaro_4d_2021,
	title = {{4D} flow {MRI}: impact of region of interest size, angulation and spatial resolution on aortic flow assessment},
	volume = {42},
	issn = {0967-3334, 1361-6579},
	shorttitle = {{4D} flow {MRI}},
	url = {https://iopscience.iop.org/article/10.1088/1361-6579/abe525},
	doi = {10.1088/1361-6579/abe525},
	language = {en},
	number = {3},
	journal = {Physiological Measurement},
	author = {Casciaro, M E and Pascaner, A F and Guilenea, F N and Alcibar, J and Gencer, U and Soulat, G and Mousseaux, E and Craiem, D},
	year = {2021},
	pages = {035004},
}

@article{zhong_intracardiac_2019,
	title = {Intracardiac {4D} {Flow} {MRI} in {Congenital} {Heart} {Disease}: {Recommendations} on {Behalf} of the {ISMRM} {Flow} \& {Motion} {Study} {Group}},
	volume = {50},
	issn = {1053-1807, 1522-2586},
	shorttitle = {Intracardiac {4D} {Flow} {MRI} in {Congenital} {Heart} {Disease}},
	url = {https://onlinelibrary.wiley.com/doi/10.1002/jmri.26858},
	doi = {10.1002/jmri.26858},
	language = {en},
	number = {3},
	journal = {Journal of Magnetic Resonance Imaging},
	author = {Zhong, Liang and Schrauben, Eric M. and Garcia, Julio and Uribe, Sergio and Grieve, Stuart M. and Elbaz, Mohammed S.M. and Barker, Alex J. and Geiger, Julia and Nordmeyer, Sarah and Marsden, Alison and Carlsson, Marcus and Tan, Ru‐San and Garg, Pankaj and Westenberg, Jos J.M. and Markl, Michael and Ebbers, Tino},
	year = {2019},
	pages = {677--681},
}

@article{raimondi_prevalence_2021,
	title = {Prevalence of {Venovenous} {Shunting} and {High}-{Output} {State} {Quantified} with {4D} {Flow} {MRI} in {Patients} with {Fontan} {Circulation}},
	volume = {3},
	issn = {2638-6135},
	url = {http://pubs.rsna.org/doi/10.1148/ryct.210161},
	doi = {10.1148/ryct.210161},
	language = {en},
	number = {6},
	journal = {Radiology: Cardiothoracic Imaging},
	author = {Raimondi, Francesca and Martins, Duarte and Coenen, Raluca and Panaioli, Elena and Khraiche, Diala and Boddaert, Nathalie and Bonnet, Damien and Atkins, Melany and El-Said, Howaida and Alshawabkeh, Laith and Hsiao, Albert},
	year = {2021},
	pages = {e210161},
}

@article{sotelo_impact_2022,
	title = {Impact of aortic arch curvature in flow haemodynamics in patients with transposition of the great arteries after arterial switch operation},
	volume = {23},
	copyright = {https://academic.oup.com/journals/pages/open\_access/funder\_policies/chorus/standard\_publication\_model},
	issn = {2047-2404, 2047-2412},
	url = {https://academic.oup.com/ehjcimaging/article/23/3/402/6124832},
	doi = {10.1093/ehjci/jeaa416},
	language = {en},
	number = {3},
	journal = {European Heart Journal - Cardiovascular Imaging},
	author = {Sotelo, Julio and Valverde, Israel and Martins, Duarte and Bonnet, Damien and Boddaert, Nathalie and Pushparajan, Kuberan and Uribe, Sergio and Raimondi, Francesca},
	year = {2022},
	pages = {402--411},
}

@article{isorni_4d_2020,
	title = {{4D} flow cardiac magnetic resonance in children and adults with congenital heart disease: {Clinical} experience in a high volume center},
	volume = {320},
	issn = {01675273},
	shorttitle = {{4D} flow cardiac magnetic resonance in children and adults with congenital heart disease},
	url = {https://linkinghub.elsevier.com/retrieve/pii/S0167527320334690},
	doi = {10.1016/j.ijcard.2020.07.021},
	language = {en},
	journal = {International Journal of Cardiology},
	author = {Isorni, Marc-Antoine and Moisson, Louis and Moussa, Nidal Ben and Monnot, Sébastien and Raimondi, Francesca and Roussin, Régine and Boet, Angèle and Van Aerschot, Isabelle and Fournier, Emmanuelle and Cohen, Sarah and Kara, Meriem and Hascoet, Sébastien},
	year = {2020},
	pages = {168--177},
}

@inproceedings{le2017computationally,
  title={Computationally efficient cardiac views projection using 3D convolutional neural networks},
  author={Le, Matthieu and Lieman-Sifry, Jesse and Lau, Felix and Sall, Sean and Hsiao, Albert and Golden, Daniel},
  booktitle={International Workshop on Deep Learning in Medical Image Analysis},
  pages={109--116},
  year={2017},
  organization={Springer}
}

@article{fan2025avp,
  title={AVP-AP: Self-supervised Automatic View Positioning in 3D cardiac CT via Atlas Prompting},
  author={Fan, Xiaolin and Wang, Yan and Zhang, Yingying and Bao, Mingkun and Jia, Bosen and Lu, Dong and Gu, Yifan and Cheng, Jian and Zhu, Haogang},
  journal={IEEE Transactions on Medical Imaging},
  year={2025},
  volume={44},
  number={7},
  doi={10.1109/TMI.2025.3554785},
  publisher={IEEE}
}

@article{rodriguez2018aortic,
  title={Aortic flow patterns and wall shear stress maps by 4D-flow cardiovascular magnetic resonance in the assessment of aortic dilatation in bicuspid aortic valve disease},
  author={Rodr{\'\i}guez-Palomares, Jos$\Omega$ Fernando and Dux-Santoy, Lydia and Guala, Andrea and Kale, Raquel and Maldonado, Giuliana and Teixid{\'o}-Tur{\`a}, Gisela and Galian, Laura and Huguet, Marina and Valente, Filipa and Gutierrez, Laura and others},
  journal={Journal of Cardiovascular Magnetic Resonance},
  volume={20},
  number={1},
  pages={28},
  year={2018},
  publisher={Elsevier}
}

@inbook{hennemuth2011fast,
  title={Fast interactive exploration of 4D MRI flow data},
  author={Hennemuth, Anja and Friman, Ola and Schumann, Christian and Bock, Jelena and Drexl, Johann and Huellebrand, Markus and Markl, Michael and Peitgen, H-O},
  booktitle={Medical Imaging 2011: Visualization, Image-Guided Procedures, and Modeling},
  volume={7964},
  publisher={SPIE},
  doi = {doi.org/10.1117/12.878202},
  pages={110--120},
  year={2011},
  organization={SPIE}
}

@article{taha2015metrics,
  title={Metrics for evaluating 3D medical image segmentation: analysis, selection, and tool},
  author={Taha, Abdel Aziz and Hanbury, Allan},
  journal={BMC medical imaging},
  volume={15},
  number={1},
  pages={29},
  year={2015},
  publisher={Springer}
}

@article{massey1951kolmogorov,
  title={The Kolmogorov-Smirnov test for goodness of fit},
  author={Massey Jr, Frank J},
  journal={Journal of the American statistical Association},
  volume={46},
  number={253},
  pages={68--78},
  year={1951},
  publisher={Taylor \& Francis}
}

@article{levene1960robust,
  title={Robust tests for equality of variances},
  author={Levene, Howard},
  journal={Contributions to probability and statistics},
  pages={278--292},
  year={1960},
  publisher={Stanford University Press}
}

@inproceedings{wobbrock2011aligned,
  title={The aligned rank transform for nonparametric factorial analyses using only anova procedures},
  author={Wobbrock, Jacob O and Findlater, Leah and Gergle, Darren and Higgins, James J},
  booktitle={Proceedings of the SIGCHI conference on human factors in computing systems},
  pages={143--146},
  year={2011}
}

@inproceedings{elkin2021aligned,
  title={An aligned rank transform procedure for multifactor contrast tests},
  author={Elkin, Lisa A and Kay, Matthew and Higgins, James J and Wobbrock, Jacob O},
  booktitle={The 34th annual ACM symposium on user interface software and technology},
  pages={754--768},
  year={2021}
}



\appendix

\setcounter{table}{0}
\renewcommand{\thetable}{A\arabic{table}}

\section{Appendix}\label{Appendix}
The 4D flow MRI datasets in this study were acquired as follows:

Thirty two scans from healthy volunteers were obtained in a clinical GE 1.5T Signa scanner (GE Healthcare, Waukesha, WI, USA) using the Vastly undersampled Isotropic Projection Reconstruction (VIPR) technique \citep{gu2005pc}. Imaging parameters included a field of view (FOV) 400 mm $\times$ 400 mm $\times$ 400 mm, voxel size of 2.5 mm $\times$ 2.5 mm $\times$ 2.5 mm, flip angle $8^\circ$, repetition time 4.2--6.4 ms, echo time 1.9--3.7 ms, and velocity encoding 200 cm/s. This dataset was reconstructed with a temporal resolution that ranged between 21--32 ms.

Eight scans from repaired aortic coarctation and 5 from tetralogy of Fallot were obtained using GE 1.5T Optima scanner (GE Healthcare, Waukesha, WI, USA). 4D-flow imaging parameters were as follows: spatial resolution (row $\times$ column $\times$ slice) = $2.34 \times 2.34 \times 2.5$ mm$^3$, temporal resolution = $37.62 \pm 4.43$ ms, flip angle = $5^\circ$; field of view (FOV) = $236 \times 226 \times 134$ mm, velocity sensitivity (Venc) = 250 cm/s, echo time = 2.66 ms, pulse repetition time = 4.78--4.80 ms.

Seventeen scans from healthy volunteers and 8 from bicuspid aortic valve patients were obtained in a 3T Prisma scanner (Siemens, Erlangen, Germany). ECG-gated 4D-flow MRI (WIP 785A) was acquired during free breathing using navigator gating of diaphragmatic motion \citep{garcia_aortic_2021}. 4D-flow imaging parameters were as follows: spatial resolution (row $\times$ column $\times$ slice) = $2.0$--$2.5 \times 2.0$--$2.5 \times 2.4$--$3.5$ mm$^3$, temporal resolution = $36.24$--$40.56$ ms, flip angle = $15^\circ$; field of view (FOV) = $240$--$350 \times 320$--$400$ mm$^2$, bandwidth = $455$--$495$ Hz/Pixel, velocity sensitivity (Venc) = $150$--$550$ cm/s, echo time = $2.01$--$2.35$ ms, pulse repetition time = $4.53$--$5.07$ ms.

Eighteen scans from healthy volunteers were obtained in a 3T Philips MRI scanner (Philips Achieva, Best, The Netherlands). 4D-flow imaging parameters were as follows: spatial resolution (row $\times$ column $\times$ slice) = $2.34 \times 2.34 \times 2.5$ mm$^3$, temporal resolution = $37.62 \pm 4.43$ ms, flip angle = $5^\circ$; field of view (FOV) = $236 \times 226 \times 134$ mm, velocity sensitivity (Venc) = 250 cm/s, echo time = 2.66 ms, pulse repetition time = 4.78--4.80 ms.



\end{document}